\documentclass[preprint,12pt,authoryear]{elsarticle}

\usepackage{amssymb}
\usepackage{amsmath}

\usepackage{bm}
\usepackage{color}
\usepackage{xcolor}
\usepackage{amsmath,amssymb,amsfonts}
\usepackage{algorithmic}
\usepackage{graphicx}
\usepackage{textcomp}
\usepackage{natbib}
\usepackage{url}
\usepackage{hyperref}
\usepackage{svg}
\usepackage{booktabs}
\usepackage{multirow}
\usepackage{soul}
\usepackage{ulem}

\setcitestyle{authoryear, open={(},close={)}}

\journal{Medical Image Analysis}

\begin{document}

\begin{frontmatter}



\title{CoMetaPNS: Continually Meta-learning \\Personalized Neural Surrogates for\\ Cardiac Electrophysiology Simulations} 

%

\author[a]{Ryan Missel}
\author[b]{Xiajun Jiang}
\author[a]{Linwei Wang}


\affiliation[a]{organization={Golisano College of 
Computing and Information Sciences, Rochester Institute of Technology},
            city={Rochester},
            state={NY},
            country={USA}}
\affiliation[b]{organization={Department of Computer Science, Rowan University},
            city={Glassboro},
            state={NJ},
            country={USA}}
\affiliation[c]{organization={The University of Utah,},
            city={Salt Lake City},
            state={UT},
            country={USA}}

\begin{abstract}
Personalized virtual heart simulations face challenges in model personalization and computational cost. While neural surrogates offer state-of-the-art solutions, they typically address either efficient personalization or training generalizable models. Recent work reframes this by learning the process of personalizing a surrogate using limited subject-specific context data, through few-shot generative modeling with set-conditioned surrogates and meta-learned amortized inference. These methods, however, assume a static and diverse training distribution with known task identifiers. When new data becomes available, they require costly retraining with all prior data to avoid catastrophic forgetting - a phenomena where the model forgets earlier tasks when trained on new ones. This is a major limitation in clinical settings where often unlabeled data arrives sequentially and full retraining is infeasible.
This paper presents a new continual meta-learning framework to achieve personalized neural surrogates able to not only continually integrate information but also identify whether incoming data stems from a known or unknown dynamics source. By leveraging a continual Bayesian Gaussian Mixture Model over a memory buffer, our framework can infer the identifiers and relationships of data over time - required for effective meta-learning. 
Empirical results on synthetic cardiac data demonstrate superior simulation forecasting, computational scalability, and resilience to catastrophic forgetting compared to existing baselines.
\end{abstract}

\begin{graphicalabstract}
  \makebox[\textwidth][c]{\includegraphics[width=1.3\textwidth]{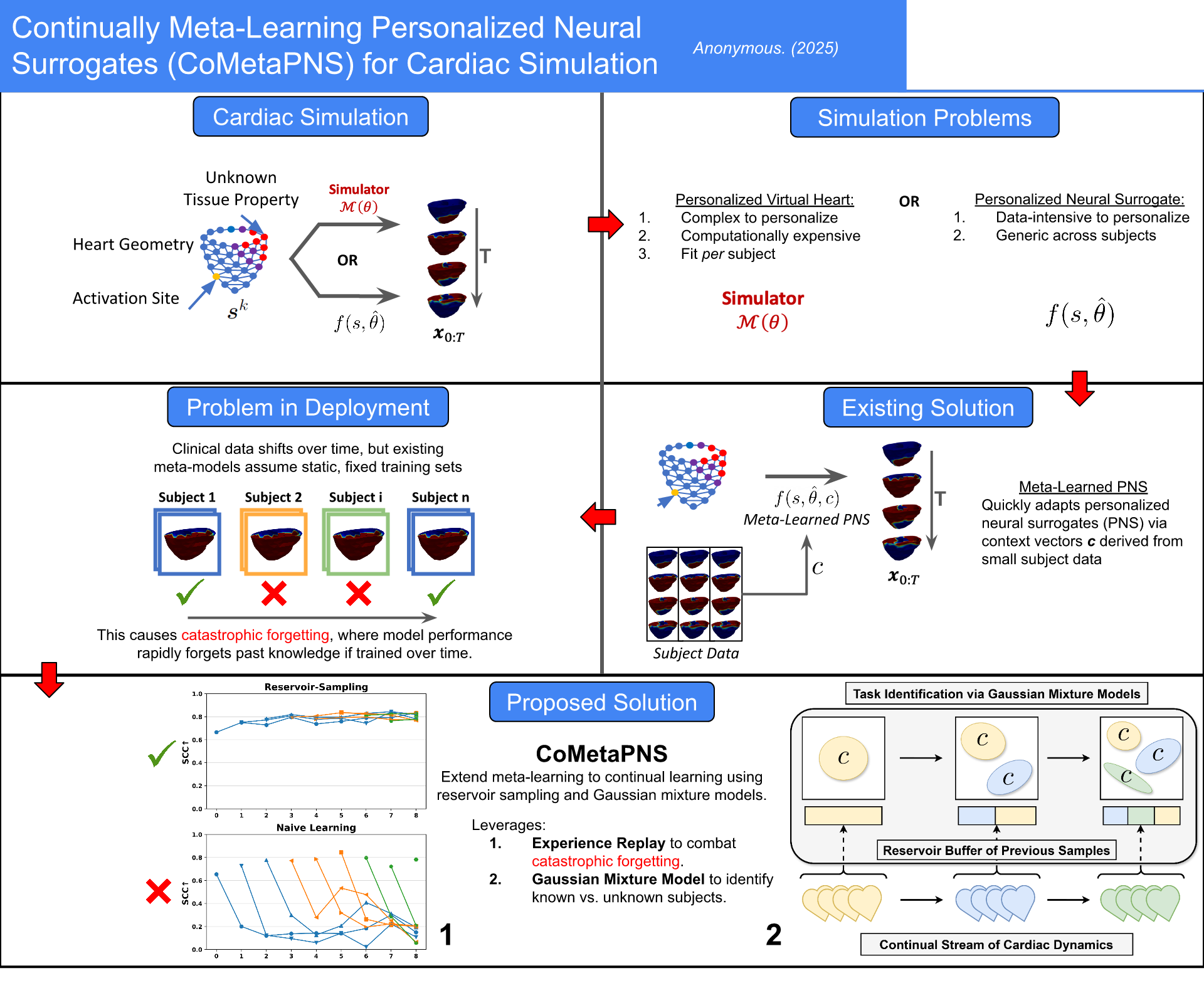}}%
\end{graphicalabstract}

\begin{highlights}
\item Personalized heart models are hard to scale due to tuning time and compute cost
\item Meta-learning quickly personalizes neural surrogates from low subject-specific data
\item Clinical data shifts over time, but existing meta-models assume fixed training sets
\item We extend meta-learning to continual learning using reservoir and clustering tools
\item Our method adapts to new subjects without retraining and avoids forgetting past data
\end{highlights}


\end{frontmatter}



\section{Introduction}
\label{sec:introduction}
Personalized virtual models of cardiac electrophysiology, tailored to the observational data of individual subjects, have demonstrated utility in clinical applications such as treatment planning \citep{prakosa2018personalized} and arrhythmia risk stratification \citep{arevalo2016arrhythmia,trayanova2024computational}. However, broader clinical integration faces two key challenges. First, calibrating these models to match an individual's physiology remains a complex and time-intensive task, particularly for parameters that cannot be directly observed, such as tissue material properties \citep{niederer2020creation,trayanova2024computational}. Second, the high computational demands of running these simulations limit their scalability, making it difficult to embed them into clinical workflows or perform thorough uncertainty analysis \citep{niederer2020creation,cluitmans2024digital}. 

There has been considerable advancement in personalizing cardiac model parameters \citep{miller2021implementation,dhamala2018quantifying,wong2015velocity,sermesant2012patient,zettinig2013fast,cluitmans2024digital}, particularly in the context of electrophysiological (EP) modeling. Early approaches typically involved iterative optimization schemes that required repeated evaluations of the computationally intensive EP simulation model \citep{wong2015velocity,sermesant2012patient}. More recent efforts have incorporated modern machine learning (ML)  techniques, including active learning \citep{dhamala2020embedding}, reinforcement learning \citep{neumann2020machine}, and direct ML of the parameter-to-observation mapping \citep{coveney2021bayesian,giffard2016noninvasive}. While these approaches have reduced the number of simulation calls required during optimization, the overall process of personalization remains complex and resource-intensive. Furthermore, even after obtaining a personalized model, the computational burden of running the simulation remains a significant obstacle for clinical-scale deployment.

Simultaneously, the field of deep learning (DL) has seen growing interest in building efficient neural approximations for computationally expensive scientific simulations \citep{Kasim20published}. However, progress on neural surrogate modeling for cardiac EP remains limited. Early results have primarily focused on 2D domains \citep{cantwell2019rethinking,kashtanova2021ep} and 3D simulations of the left atrium \citep{fresca2021pod}, with more recent work extending to full-field simulations of the both ventricles \citep{martinez2025full} or whole-heart electromechanical simulations \citep{salvador2024whole}.
A central challenge in this space is the dependence of simulations on physiological parameters, denoted $\theta$, such as material properties. Most approaches aim to learn a single neural function $f(\theta)$ to approximate the behavior of a simulation model $\mathcal{M}(\theta)$. This introduces two main issues. First, learning an accurate $f(\theta)$ requires a large dataset of paired samples $\{ \theta_i, \mathcal{M}(\theta_i) \}$ simulated across the input space of $\theta$, which is computationally challenging to generate. While the emergence of physics-informed neural networks (PINNs)  alleviates this need of data by supervising the neural networks with the analytical partial differential equations (PDEs) of $\mathcal{M}(\theta)$,  it requires a distinct PINN to be trained from scratch for each particular value $\theta_i$  (and boundary/initial conditions for that matter)  \citep{herrero2022ep,gomez2025simulation}.  In general, in either approach,  there is the difficulty  of ensuring sufficient coverage of the parameter space during training,  such that the trained surrogate is robust to out-of-distribution parameter values at practical deployment.
Second, even if $f(\theta)$ is well-approximated, its application depends on having accurate patient-specific values of $\theta$, which are often unknown or unobservable, in order to become personalized \citep{trayanova2024computational}.

Recently, a \textit{learning-to-learn} approach has been proposed to 
address the challenges above associated with learning how to personalize a neural surrogate itself \citep{jiang2022few}. The guiding principle is being interested in a set of personalized neural functions, rather than one single generic neural function, as the simulation surrogate: therefore, instead of \textit{learning a single neural surrogate,} one can \textit{learn the process of learning} a personalized neural surrogate from a small number of available data of a subject (\textit{context} data). This is cast as a novel formulation of few-shot generative modeling via Bayesian meta-learning \citep{jiang2022few}.
It includes two elements: 1) a set-conditioned generative model as the neural surrogate for cardiac EP simulation that, conditioned on an abstract latent embedding that represents patient-specific information, learns to generate \textit{target} simulations that are personalized to an individual, and 2) a meta-model of amortized variational inference (VI) that learns to extract such patient-specific embedding via feed-forward embedding from context data of \textit{variable} sizes from a patient. Compared to optimization-based meta-learning methods \citep{finn2017model,ravi2016optimization}, this type of feed-forward meta-models remove the need of further training or fine-tuning, obtaining a model at meta-test time via simple feed-forward embedding of context data. With this, at test time, a personalized neural surrogate can be quickly obtained by simple feed-forward embedding of a small and flexible number of data available from a subject. 

Despite significant advancements in neural surrogates and their personalization - particularly through meta-learning approaches - these methods share a fundamental limitation: the assumption of access to a \textit{static} training distribution of \textit{diverse} subjects. When fast adaptation to a subject within or \textit{near} the training distribution is possible, 
if data from \textit{novel} subjects becomes available, such models require expensive retraining with the entirety of previous data to avoid \textit{catastrophic forgetting} - the well-known phenomenon where models forget earlier tasks when trained only on new information \citep{kirkpatrick2017overcoming}. In clinical practice, subject data typically arrives sequentially and often exhibits distributional shifts, rendering models trained on prior cohorts insufficient for generalization to new subjects or even known subjects changing over time \citep{kolk2024dynamic}. Consequently, there is a critical need within the clinical workflow for efficient simulation surrogates that can be rapidly personalized to individual subjects while maintaining performance across non-stationary distributions of cardiac data \citep{bhagirath2024bits}.

The concept of continual meta-learning (CML), which combines 
continual training with few‑shot adaptation, is well-suited for addressing this challenge. Existing CML methods, however, face two core limitations in this context. Most works rely on gradient-based meta-learners like model-agnostic meta-learning (MAML) \citep{MAML}, which leverages a shared parameter initialization to adapt to \textit{context} data in a few gradient-update steps. This requires computationally expensive test-time fine-tuning and can fail to generalize beyond image classification and low-dimensional regression tasks \citep{SNAIL}. 
Second, most employ continual strategies that are task-agnostic, leveraging the approximate equivalence between continual and meta-learning objectives in aligning gradients by directly applying the meta-model adapted on actively-streaming data to all prior tasks \citep{MER,LAMAML}. These approaches fundamentally conflict with the requirements of mesh-based cardiac data, where per-vertex spatial variations induce patient-specific dynamics (that a shared initialization cannot accommodate), and where clinical deployment demands low-latency inference.
Addressing these gaps requires moving beyond gradient-based and task-agnostic CML methods toward approaches that infer task identities directly and can adapt without expensive gradient updates.

In this paper, we present a novel concept to achieve \textit{continually-adapted personalized neural surrogates} in a single coherent end-to-end framework of continual meta-learning (CoMetaPNS), addressing the challenge of continual adaptation central to personalized cardiac modeling. Specifically, we develop a continual Bayesian meta-learning framework that (1) learns to learn a personalized neural surrogate from limited, variable-sized subject-specific observations; (2) incorporates a task-relational past-sample reservoir that uses continual Bayesian Gaussian Mixture Models to distinguish whether new data stems from previously encountered or novel dynamics; and (3) continually updates the meta-inference model to preserve past knowledge while integrating new information, thereby avoiding catastrophic forgetting. By enabling automatic task identification and directly inferring per-vertex conditioning variables, CoMetaPNS aligns with the spatial variability inherent in patient-specific cardiac meshes, advancing beyond the limitations of existing gradient-based and task-agnostic CML approaches and
enabling continual and rapid personalization  even under sequential and non-stationary data streams typical of clinical deployment scenarios.

We evaluated CoMetaPNS on synthetic non-stationary cardiac data streams and its generalization to real cardiac data, highlighting the joint necessity of its meta-learning and continual learning components. We included comparisons to the following baselines: (1) individually-personalized cardiac simulation models optimized using established techniques \citep{dhamala2018high}, 
2) generic neural surrogates lacking set conditioning or meta-inference, and (3) alternative meta-inference approaches representing standard practices in continual meta-learning (CML) literature. CoMetaPNS achieved superior personalization and predictive error at significantly reduced computational cost compared to both conventionally-optimized simulations and existing meta-inference baselines. Importantly, we found that neither meta-learning nor continual learning alone sufficed under non-stationary data conditions - only their integration provided stable performance. Additionally, we demonstrated the utility of the feed-forward embedding alongside a continual Bayesian Gaussian mixture model, which reliably distinguished between re-emerged (\textit{i.e.,} known) and novel data sources. This enabled control over updating the meta-model's parameters or applying fast adaptation alone.

\section{Problem Formulation}

 Consider a cardiac electrophysiology simulation model \( x_{1:T} = \mathcal{M}(v; \theta) \), where $v$ is a known input (e.g., electrical stimulation applied to the virtual heart), \( \theta \) is an unknown parameter representing subject-specific properties (e.g., tissue conductivity), and $x_{1:T}$ are heart surface unipolar potential maps.
The objective of model personalization is to estimate \( \hat{\theta} \) that minimizes the discrepancy between observed and simulated \( x_{1:T} \). A personalized neural surrogate aims to learn a function \( f(v, \hat{\theta}) \) that both 1) approximates the simulation output \( \mathcal{M}(v; \hat{\theta}) \) for computational efficiency and 2) aligns with the individual's observations for personalization.

A natural approach is to learn a neural function \( f(v, \theta) \) that approximates the simulation model \( \mathcal{M}(v; \theta) \) across the space of \( \theta \) \citep{kashtanova2021ep, fresca2021pod}, yielding a generic surrogate that either requires an explicit input of \( \theta \) or a separate optimization step for personalization. In contrast, leveraging state-of-the-art meta-learning, we aim to derive a set of personalized neural surrogates that automatically adapt to a small, variable-sized set of context observations 
\( \mathcal{X}^s = \{ x_{1:T}^{s,i} \}_{i=1}^{\kappa} \) 
from an individual subject, 
where $\kappa$ is the number of observations available for the subject. 
This is formulated in the framework of Bayesian meta-learning as:
\[
p(\hat{x}_{1:T} | v, \mathcal{X}^s) = \int p(\hat{x}_{1:T} | v, c) \, q_{\zeta}(c | \mathcal{X}^s) \, dc,
\]
where \( p(\hat{x}_{1:T} | v, \mathcal{X}^s) \) describes the likelihood of the sequence $\hat{x}_{1:T}$ given known input $v$ and context observation sequences $\mathcal{X}^s$.
The generative model \( p(\hat{x}_{1:T} | v, c) \) conditions on known stimulation input \( v \) and a latent embedding \( c \) personalized to the subject, while \( q_{\zeta}(c | \mathcal{X}^s) \) is a meta-model parameterized by \( \zeta \) that learns to extract patient-specific \( c \) through feed-forward embedding of the context observations \( \mathcal{X}^s \).

We aim to maintain the performance of the meta-model over a non-stationary stream of observations as data from different subjects become available over time, which consists of varying heart meshes and material properties $\theta$. To reflect practical constraints in clinical settings, we make the following assumptions regarding the nature and availability of such data:
\begin{enumerate}
    \item We do not assume access to the complete historical data distribution as it is often infeasible due to memory/storage limitations and privacy concerns. Instead, we assume access to a fixed-size memory buffer that can hold up to $\mathcal{M}$ samples.
    \item We assume that we know when the data source - \textit{i.e.}, the subject - changes, a reasonable assumption within clinical workflows. We however do not assume access to ground-truth identifiers for these sources: thus, we do not know the similarity or dissimilarity about the underlying conditions among subjects; similarly,
    we do not know whether any newly-presented data corresponds to a previously-known or novel subject.
    \item We assume the presence of \textit{local stationarity} where data from each subject are presented for some period of time. This is a common assumption utilized in continual-learning literature to allow the neural model to optimize sufficiently and stably before a new data source has the potential of appearing.
\end{enumerate}
Within these constraints, our goal is to effectively manage the limited memory buffer to approximate the overall data distribution over time and infer the relationship between incoming and previously encountered data sources, 
such that we can learn to adapt our neural surrogates to data from such non-stationary distributions of subjects without forgetting.

\section{Methodology}
\begin{figure*}[!t]
    \centering
    \includegraphics[width=\textwidth]{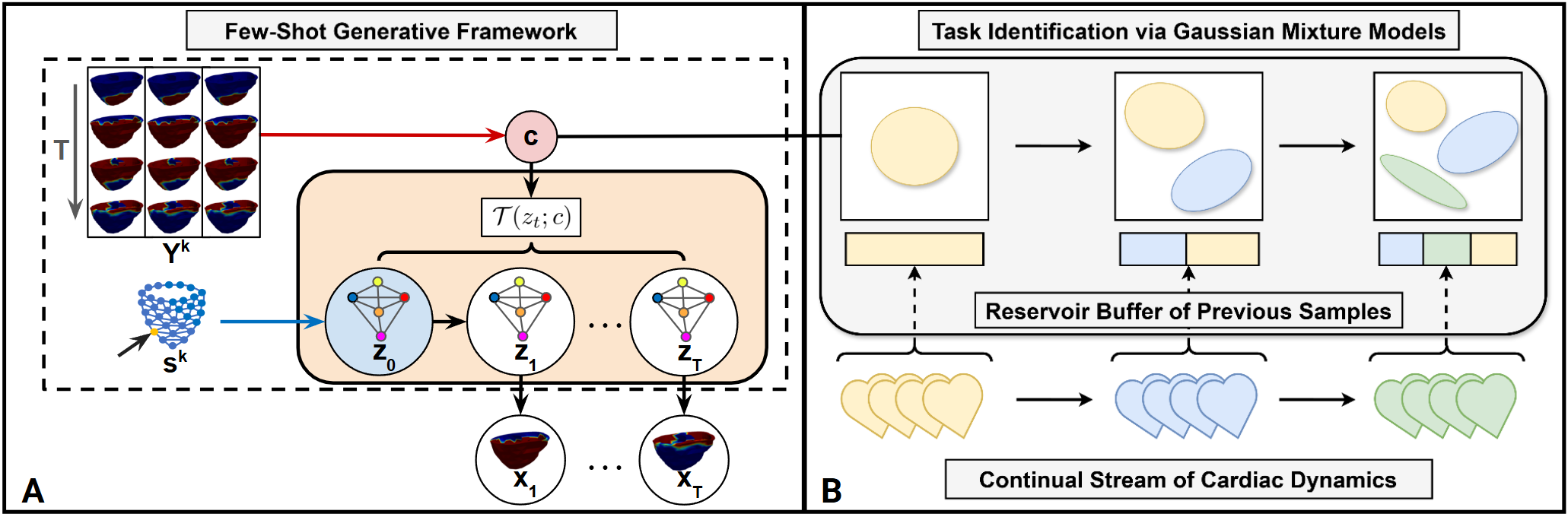}
    \caption{Overview of \textit{CoMetaPNS}, showing A) the framework of few-shot generative modeling via Bayesian meta-learning that B) continually aggregate a heterogeneous data stream of cardiac electrophysiology dynamics with a sample reservoir that identifies subjects via Gaussian mixture models.}
    \label{overview}
\end{figure*} 

Fig.~\ref{overview} gives an overview of CoMetaPNS. It includes two major components: A) an adaptive neural surrogate realized via feed-forward meta-learning, and B) a continual learning strategy for learning this meta-model across continuous data streams of diverse cardiac EP dynamics.

\subsection{Adaptive Neural Surrogates via Feedforward Meta-Learning}
The adaptive neural surrogate is similar to that described in \citep{jiang2022few}, which includes a set-conditioned generative model 
(Section \ref{subsubsec:generative}) 
and a meta-inference strategy 
for extract patient-specific embedding to condition this generative model 
(Section \ref{subsubsec:inference}).

\subsubsection{Set-Conditioned Generative Model}
\label{subsubsec:generative}
The generative neural surrogate is conditioned on a context-set embedding \(c \) and known stimulation input $v$. It consists of two components: a temporal transition model \( \mathcal{T}_{\theta_t} \) that governs the evolution of the latent state \( z_t \), and a spatial emission model \( \mathcal{G}_{\theta_s}  \) that maps the latent state to the high-dimensional cardiac mesh output \( x_t \). These components are defined as:
\[
\begin{aligned}
\text{Transition:} \quad & z_{t+1} = \mathcal{T}_{\theta_t} (z_t), \\
\text{Emission:} \quad & \hat{x}_{t+1} = \mathcal{G}_{\theta_s} (z_{t+1}).
\end{aligned}
\]
The initial latent state \( z_0 \) is derived from the known stimulation \( v \) using a neural function composed of a linear layer applied to the embedding of \( v \), denoted as \( z_0 = f_{\rho}(v) \). 

\textbf{Temporal transition $\mathcal{T}_{\theta_t}$:}
The temporal transition function $\mathcal{T}_{\theta_t}$ is inspired by Gated Recurrent Units (GRUs) \citep{chung2014empirical}.
Note that, with regular GRUs, $\mathcal{T}_{\theta_t} $ would be global to the training data rather than subject-specific. Instead, we condition $\mathcal{T}_{\theta_t} $ on 
the context-set embedding by creating conditional gated transition functions:
\begin{equation}
    \begin{aligned}
    \quad z_t^{(1)} &= \mathrm{ELU}(\alpha_1 z_t + \beta_1 c + \gamma_1), \\
    \quad z_t^{(2)} &= \mathrm{ELU}(\alpha_2 z_t + \beta_2 c + \gamma_2), \\
    \quad z_t^{(3)} &= \alpha_3 z_t + \beta_3 c + \gamma_3, \\
    g_t &= \sigma(W_1 z_t^{(1)} + b_1), \\
    h_t &= \mathrm{ELU}(W_2 z_t^{(2)} + b_2), \\
    z_{t+1} &= (1 - g_t) \odot (W_3 z_t^{(3)} + b_3) + g_t \odot h_t,
    \end{aligned}
\end{equation}
where $c$ $\sim p(c| \mathcal{X}^s)$ (see Section \ref{subsubsec:inference}) and $\theta_t =\{W_i, b_i, \alpha_i, \beta_i, \gamma_i \}_{i=1}^3$ are learnable parameters. The model has flexibility to choose a linear transition for some dimensions and non-linear transition for the others. 

\textbf{Spatial emission $\mathcal{G}_{\theta_s}$:}
Since \( x \) resides on the 3D geometry of the heart, we implement the emission function \( \mathcal{G}_{\theta_s} \) using Graph Convolutional Neural Networks (GCNNs). We represent the heart’s triangular mesh as an undirected graph, where edges between vertices are assigned attributes defined by the normalized differences in their 3D coordinates, provided an edge exists. Encoding and decoding are performed over hierarchical representations of this geometry, constructed via a specialized mesh coarsening algorithm \citep{cgal:c-tsms-07}. To enable spatial convolution across graphs, we adopt a continuous spline kernel \citep{fey2018splinecnn}. To enhance the model’s expressivity and depth, we incorporate residual blocks using skip connections implemented with 1D convolutions, following \citep{jiang2020learning}.

\subsubsection{Meta-Model for Amortized Variational Inference}
\label{subsubsec:inference}
We introduce a meta-model $p_{\phi}(c | \mathcal{X}^s)$ to model the conditional distribution of c given a context set $\mathcal{X}^s$, realized via a feed-forward neural network $h_\phi$. First, each sample $x^s_{1:T} \in \mathcal{X}^s$ is embedded through the neural function $h_{\phi}(x^s_{1:T})$ that uses a GCN-GRU cell \citep{jiang2021label} to obtain the sequential information from the graph, and aggregate it across time with a linear layer. We then average all latent embedding in $\mathcal{X}^s_k$:
\begin{equation}
    \begin{aligned}
    \frac{1}{|\mathcal{X}^s|} \sum\nolimits_{x^s_{1:T} \in \mathcal{X}^s} h_{\phi}(x^s_{1:T}),
    \end{aligned}
\end{equation}
which then parameterizes $p_{\phi} = \mathcal{N}(\bm{\mu}_c, \bm{\sigma}_c^2)$ via two separate linear layers. The conditional factor $c$ is then sampled by $c = \bm{\mu}_c + \bm{\epsilon} \odot \bm{\sigma}_c$, where $\bm{\epsilon} \sim \mathcal{N}(0, \mathbf{I})$ \citep{kingma2013auto}.

Now consider a set $\mathcal{Q}$ of subjects $\{ Q_k \}_{k=1}^{K}$, 
where the data distribution of each subject $Q_k$ is defined by $p(Q_k)$ and the distribution of subjects by $p(\mathcal{Q})$.
For each subject $Q_k$, we define two associated sets of high-dimensional cardiac data: 
1) the \textit{query} set $\mathcal{X}_k^q = \{ x_{1:T}^{q,1}, x_{1:T}^{q,2},..., x_{1:T}^{q,N_k} \}$, for which only the stimulation inputs $\mathcal{V}_k = \{ v^{1}, v^{2},..., v^{N_k} \}$ are available to forecast from; 
and 2) the \textit{context} data $\mathcal{X}_k^s = \{ x_{1:T}^{s,1}, x_{1:T}^{s,2},..., x_{1:T}^{s,M_k} \}$, consisting of fully-observed sequences used to infer the subject-specific conditioning variable $c$. Note that $M_k \ll N_k$ and, as discussed below, these sets are mutable during episodic training. 

The evidence lower bound (ELBO) we optimize for each $Q_k$ can be formulated as:
\begin{equation}
\label{eqn:metaloss_partial}
    \begin{aligned}
\sum_{i=1}^{N_k}I(x^{q,i}_{1:T},Q_k) \log p(\hat{x}^{q,i}_{1:T} | v^{k,i}, \mathcal{X}^s_k) &\geq  
    \\&\sum_{i=1}^{N_k}I(x^{q,i}_{1:T},Q_k) 
    \Big[\mathbb{E}_{q_{\zeta}(c^k | \mathcal{X}^s_k\cup x^{q,i}_{1:T})}  [ \log p(\hat{x}^{q,i}_{1:T} | c^k, v^{k,i}) ] \\ &- \mathrm{KL} ( q_{\phi}(c^k | \mathcal{X}^s_k \cup x^{q,i}_{1:T}) || p_\phi(c^k | \mathcal{X}^s_k) ) \Big] 
    \end{aligned}
\end{equation}
where $I(x_{1:T}, Q_k)$ a binary membership function that equal to $1$ if $x_{1:T}$ is in $Q_k$ and $0$ otherwise.
Note that we let $p_\phi(c^k | \mathcal{X}^s_k)$ and $q_\phi(c^k | \mathcal{X}^s_k \cup x^q_{1:T})$ share the same meta set-embedding networks to parameterize their means and variances. We further regularize $p_\theta(c^k | \mathcal{X}^s_k)$ to be close to a standard Gaussian distribution $\mathcal{N}(0, I)$,
giving a loss function $\mathcal{L}_{Q_k}(I_k)$ per $Q_k$ as:
\begin{equation}
\label{eqn:metaloss}
    \begin{aligned}
        \mathcal{L}_{Q_k}(I_k) 
        & = 
    \sum_{i=1}^{N_k}
    I_k
\Big[\mathbb{E}_{q_{\phi}(c^k | \mathcal{X}^s_k\cup x^{q,i}_{1:T})} \left [ \log p(\hat{x}^{q,i}_{1:T} | c^k, v^{k,i}) \right] \\ 
    &- \lambda_1 \mathrm{KL} \left( q_{\phi}(c^k | \mathcal{X}^s_k \cup x^{q,i}_{1:T}) || p_\phi(c^k | \mathcal{X}^s_k) \right) \\ 
    &- \lambda_2 \mathrm{KL} \left( 
    p_\phi(c^k| \mathcal{X}^s_k) || \mathcal{N}(0, I) \right) \Big]
    \end{aligned}
\end{equation}
where $I_k$ is shorthand for $I(x^{q,i}_{1:T}, Q_k)$, 
and 
$\lambda_1$ and $\lambda_2$ are regularization multipliers. 
Let $\Theta = \{ \theta_s, \theta_t, \rho, \phi\}$ including all unknown weight parameters of our model described above, 
we have the overall optimization objective across $\mathcal{T}$ as:
\begin{equation}
\label{eqn:metaloss_total}
  \begin{aligned}
  \hat{\Theta}  =  \max_\Theta
\sum\nolimits_{Q_k \in \mathcal{Q}, \mathcal{Q} \sim p(\mathcal{Q})}
        \mathcal{L}_{Q_k}(I_k) 
    \end{aligned}
\end{equation}

The optimization problem in Equation \ref{eqn:metaloss_total} is solved in episodic training. In each training episode across all subjects, the input data is divided into two separate sets: a \textit{context} set $\mathcal{X}^s_k$ consists of small sets of samples from each subject and the \textit{query} set $\mathcal{X}^q$ formed by the remaining data. The model is asked to take $\mathcal{X}^s_k$ for each subject $k$, derive $c^k$, and generate samples $\mathcal{\hat{X}}^q_k$ from the target set stimulation inputs $\mathcal{V}_k$.

\subsubsection{Use of Feed-Forward Adaptation}
A natural question arises regarding the use of a feed-forward meta-learner to perform the adaptation within this setting rather than standard gradient-based meta-learners. The most common realization in CML is via the model-agnostic meta-learner (MAML) \citep{MAML}, in which the meta-model is adapted to new tasks by optimizing a learned parameter initialization using a few gradient steps on the available context data. While effective in classification tasks, gradient-based meta-learners like MAML often struggle to adapt to the optimization landscapes found in other domains \citep{SNAIL}. Moreover, in the continual environment, they require fine-tuning for every new task, including previously encountered ones, making them sensitive to optimization hyperparameters and computationally inefficient. This inefficiency becomes particularly problematic in cardiac simulation scenarios, where the underlying neural surrogate models are often inherently computationally expensive to update via backpropagation. 

Furthermore, the algorithmic prior of gradient-based meta-learners - the mechanism through which they adapt to task-specific information - conflicts with the demands of data represented as heterogeneous graph structures. In our feed-forward conditioning framework, per-vertex conditioning vectors can be generated in a way that remains agnostic to the global graph topology while still providing spatially localized adaptation when passed through a shared dynamics function. In contrast, gradient-based methods require a shared initialization across all tasks, adapted via a small number of gradient steps. In graph-based domains, where the structure varies between tasks, this shared initialization cannot capture per-vertex nuances, since vertex-specific parameters are not transferable across different graph topologies. Consequently, adaptation is restricted to the shared parameters of the dynamics function, making task-specific adaptation substantially more challenging. Within just a few gradient updates, the meta-learner must simultaneously adapt the shared dynamics to both the temporal behavior and the diverse spatial configurations of each new graph - an inherently difficult objective given the complexity of the domain.

\subsection{Continual Learning: How to Adapt Without Forgetting}
Now consider a condition where the set of subjects ${Q_k}$ and their corresponding data are presented sequentially over time. Further consider that a subject's data may reappear multiple times in this stream of data. To be general, we assume that we know when then source of the data (\textit{i.e.}, subjects) changes, but we do not know the identity of the source (\textit{i.e.}, we do not know if a newly-presented subject corresponds to a new or previously seen subject). In continual learning, this corresponds to a setting where task-boundary (\textit{i.e.}, the boundary of data-distribution switching) is known, but the identifier of the underlying task is unknown.

The learning objective as defined in Equations~\ref{eqn:metaloss_total} is optimized over a stationary distribution of $p(\mathcal{Q})$, as in standard meta-learning practice. 
As such, despite its ability to infer context and personalize the resulting neural surrogate, its meta-components become susceptible to catastrophic forgetting in non-stationary distributions of $p(\mathcal{Q})$. 
Additionally, how to properly identify the underlying task to enable context-query data pairing within the task becomes complicated. Below, we describe mechanisms to approximate the stationary subject distribution $p(\mathcal{Q})$ and the subject membership function $I_k$ in order to achieve a continual approximation of the learning objective in Equation~\ref{eqn:metaloss_total}. These include a reservoir sampling based experience replay strategy (Section \ref{subsubsec:RS}), combined with two alternative mechanisms of continual meta-learning depending on how task identifiers are estimated and their relations modeled.
(Sections \ref{sec:CatastrophicForgetting} -- \ref{subsec:task-relational}).

\subsubsection{Experience Replay based on Reservoir Sampling}
\label{subsubsec:RS}
In the non-stationary setting where only the data of one distribution is actively streaming in at a given time, a first question is how to obtain and aggregate errors from prior data distributions to update the meta-weights. For this, we adopt the reservoir sampling method commonly used in existing CML approaches, where a simple algorithm tracks the number of samples ($N$) seen and, for each incoming sample, overwrites an existing buffer sample with probability $M/N$ where $M$ is the size of the reservoir. 

With actively streaming data and replayed samples in the reservoir buffer, 
the next question is how to accurately pair context and query samples from the 
same task in order to approximate the meta-objectives as described in Equation \eqref{eqn:metaloss_total}.
Under the assumption of \textit{local stationarity}, this question is trivial for the 
streaming data which is assumed to be the current task:
as adopted in prior continual meta-learning (CML) works \citep{OSAKA,MetaBGD}, the most recent \( k \) observations are used as the context set for the current task. 
For samples in the reservoir buffer, we consider two mechanisms for estimating their task identifiers and model the relations among the identified tasks. 
Depending on the mechanisms used to estimate the task identifiers, 
we also adjust the reservoir samples to approximate $p(\mathcal{Q})$ differently which will be detailed in the next two sections.

\subsubsection{Task-Aware Meta-Learning} \label{sec:CatastrophicForgetting}
A simple mechanism is to directly leverage the known boundaries of subject switching, and assume that the incoming data source at every boundary belongs to a unique novel task. Under this assumption, we can simply use a boundary counter to assign pseudo task labels to data samples.
When sampling from the reservoir, 
context-query pairs are easily matched by using each sample's pseudo task label. 
Because we know the boundary of subject switching but not the identifiers of the subjects,
in relation to the original meta-learning objective in Equation \eqref{eqn:metaloss_total}, 
this approach may represent a subject $Q_k$ with one or multiple tasks, where each task will correspond to at most one subject, \textit{i.e.,} $Q_k = \{ T_{k,i}\}_{i=1}^{m_k}$, where $m_k$ represents the number of times the subject reappeared in the data stream. The membership function $I_{k,i} = I(x_{1:T}, T_{k,i})$ for a sample $x_{1:T}$ in task $T_{k,i}$ is exact given the known task boundary. In other words, in this approach, the meta-loss $\mathcal{L}_{Q_k}(I_k)$ per subject $Q_k$ in Equation \eqref{eqn:metaloss_total} will be approximated as:
\begin{equation}
    \label{eqn:metaloss_total}
    \begin{aligned}
        \mathcal{L}_{Q_k}(I_k) \simeq 
        \sum\nolimits_{i=1}^{m_k}
        \mathcal{L}_{T_{k,i}}(I_{k,i}) 
    \end{aligned}
\end{equation}
and the overall meta-loss over the streaming and replayed samples becomes:
\begin{equation}
    \label{eqn:metaloss_approx1}
    \begin{aligned}
        \sum\nolimits_{Q_k \in \mathcal{R}, \mathcal{R} \sim p(\mathcal{R})}
        \mathcal{L}_{Q_k}(I_k)
    \end{aligned}
\end{equation}
where 
$\mathcal{R}$ includes limited samples of each $Q_k$ in the reservoir, and
$p(\mathcal{R})$ approximates $p(\mathcal{Q})$ in the reservoir. 

This approach has the advantage of being straightforward and easy to implement. Furthermore, there is no error in estimating the task membership of a sample, which will avoid pairing context and query samples from different data distributions. However, by overlooking the underlying relationships among tasks, this simplified approach is likely to consider data
from similar and even re-appeared subjects as distinct tasks. In clinical settings, this can result in unnecessary retraining on familiar subjects instead of utilizing existing meta-knowledge, leading to wasted time and computational effort. Furthermore, as the number of subject-switching boundary increases continually, this method can result in an increasingly-fragmented reservoir buffer. Given the finite size of the reservoir, each task will be represented by a reducing number of samples: this can result in overfit to limited context-query pairs per task and, when the data distribution of some subjects are very different from the rest, they will become under-represented tasks in the reservoir and risk being rapidly forgotten. Collectively, this will result in an increasingly inaccurate $p(\mathcal{R})$ as an approximation of $p(\mathcal{Q})$ in Equation \eqref{eqn:metaloss_approx1} as the number of tasks increase with the number subject-switching boundaries.

\subsubsection{Task-Relational Meta-Learning}
\label{subsec:task-relational}

An alternative mechanism is to explicitly model task relations based on similarities among the streaming and replayed data samples. To this end, we leverage a \textit{continual} Bayesian Gaussian mixture model (GMM) to describe 
the context-embeddings from the reservoir samples as a distribution of clusters $p(\mathcal{R})$, where each cluster is defined as $\mathcal{R}_k$, within the reservoir over time and determines a sample's relation to their data distributions $p(\mathcal{R}_k)$.The GMM is updated continually by re-initializing its parameters using the cluster components and weights from the previous task, ensuring temporal consistency in clustering. This approach removes the dependence on traditional reservoir sampling strategies for memory updates; instead, the GMM itself determines which samples to retain or overwrite based on inferred task similarity.

The continual GMM integration proceeds in four stages, assuming we are at a newly encountered boundary at the $j$-th subject switch. At this point, an initial batch of new, unlabeled samples $[\mathcal{X}_k^q, \mathcal{X}_k^s]$ is observed, and the GMM-based procedure for clustering and memory update begins as follows:

\textit{i)} Using the previously fit GMM, we compute the embeddings $c_k^s$ of the current batch's \textit{context} set $\mathcal{X}_k^s$ and evaluate their per-cluster log-likelihoods. This provides a quantitative measure of whether the new data distribution aligns with previously observed sources. If the average log-likelihood of these samples is negative, we classify the data source as novel and proceed by updating the meta-model's parameters through full backpropagation. Novel data sources are assigned a unique task ID, incremented based on the number of unique tasks identified so far. Conversely, if the average log-likelihood is positive, we classify the source as known and opt for fast feed-forward adaptation without parameter updates. To assign the task ID for this known data source, we identify the cluster to which its samples belong, then assign it the majority task ID from the reservoir $C_j^{\text{reservoir}}$'s samples in that cluster. Note that we do not use the clusters themselves as task identifiers, as this could lead to per-sample misclassifications if a sample from a given data source is incorrectly clustered. Instead, we aim to maintain stable task ID assignments over time by grouping data sources consistently, while still leveraging the GMM's cluster likelihoods to perform effective novelty classification. Overall, this identification mechanism enables dynamic control over when to apply efficient adaptation versus full meta-model training, based on inferred novelty, as further demonstrated in Section~\ref{sec:ClusterAblation}.

\textit{ii)} During the course of task $j$, a subset of its streaming samples is stored in an auxiliary memory buffer $C_j^{\text{active}}$ using the standard reservoir sampling procedure. This buffer serves as a temporary holding set and is maintained independently of whether the current task has been identified as novel or known. The size of $C_j^{\text{active}}$ is strictly smaller than that of the main reservoir and can be dynamically adjusted as a fraction of the current number of clusters, specifically $\frac{1}{||\mathcal{R}||}$, to ensure balanced representation as the number of identified sources grows.

\textit{iii)} When the $(j+1)$-th subject-switching boundary is encountered, prior to processing any new incoming data, the GMM is re-fitted using the context embeddings from both the existing reservoir samples $C_j^{\text{reservoir}}$ and the auxiliary buffer $C_j^{\text{active}}$ collected during step \textit{ii}. This re-fitting occurs regardless of the task classification outcome in step \textit{i} (novel or known). Specifically, a new mixture component is initialized from the mean and covariance of $C_j^{\text{active}}$, and assigned a low initial weight. Upon re-fitting, the updated weight of this component reflects the novelty of the data: high weight indicates a novel source not yet represented in the reservoir, while low weight suggests redundancy with existing clusters. To prevent unbounded growth of the mixture model, components with weights below $0.05$ are pruned. 

\textit{iv)} After fitting the GMM, but still prior to new data processing, the next step is to integrate the auxiliary buffer's samples into $C_j^{\text{reservoir}}$ while ensuring two objectives: 1) preserving the reservoir buffer size and 2) maintaining overall task distributional balance. This integration strategy is guided by the novelty classification from step \textit{i}. If the task is classified as novel, we sample an equal number of entries per unique task from the union of the reservoir $C_j^{\text{reservoir}}$ and auxiliary reservoir $C_j^{\text{active}}$, ensuring balanced representation across all tasks. If the task is identified as known, we restrict the update to rebalancing only the corresponding task’s entries within the reservoir, allowing new samples to be integrated without disrupting the broader distributional structure.

Finally, in relation to the original meta-learning objective in Equation \eqref{eqn:metaloss}, 
this approaches treats each GMM cluster $\mathcal{R}_k$ as a unique task,
assigning each cluster its own pseudo-label for its assigned reservoir samples that spans across subject-switching boundaries. Given this is a data-driven clustering process, it is possible that a cluster may consist of data samples stemming from more than one underlying subject. This is a beneficial aspect of the clustering in which subjects whose underlying data distributions overlaps significantly are mapped to a similar context-embedding space. This allows the meta-model to consider the relationship among subjects and prevent unnecessary retraining on similar data distributions. Formally, the data distribution as described by the GMM within the reservoir buffer, 
including the distribution $p(R_k)$ of each cluster $R_k$ and the distribution of the clusters $p(\mathcal{R})$, approximate the actual data distribution of each subject $p(Q_k)$ and subject distribution $p(\mathcal{Q})$, respectively. 
This gives a continual approximation of the meta-objective in Equation \eqref{eqn:metaloss_total} as:
\begin{equation}
\label{eqn:metaloss_approx1}
  \begin{aligned}
\sum\nolimits_{
R_k\in \mathcal{R},
\mathcal{R} \sim p(\mathcal{R})}
\mathcal{L}_{R_{k}}(M_{k}) 
    \end{aligned}
\end{equation}
where $\mathcal{L}_{R_{k}}(M_{k})$ is calculated over the distribution of $p(R_k)$ and the estimated cluster assignment $M_k$ approximates the true subject membership $I_k$.

\section{Synthetic-Data Experiments}
\label{sec:exp:synthetic}

In all experiments, CoMetaPNS consisted of 4 GCNN blocks and 2 regular convolution layers in the encoders, 1 context-set aggregator with a GCN-GRU block followed by a linear layer to compress time and another linear layer for feature extraction, 1 conditional gated transition unit for the generation of latent dynamics, and 4 GCNN blocks and 2 regular convolution layers in the decoder. We used Adam optimizer \citep{kingma2014adam}. The learning rate at the beginning of each task was set at $1 \times 10^{-3}$ with a learning rate decreasing at a rate of 0.5 every 100 iterations. The two KL multipliers were: $\lambda_1 = 10^{-4}$ and $\lambda_2 = 0.1$. All experiments were run on Tesla T4s with 16GB memory. Our implementation is available upon acceptance.

\subsection{Experimental Setup}
\subsubsection{Data}
In synthetic data, we considered 3 heart meshes comprising 448, 475, and 480 nodes, respectively, with a combination of 12 distinct spatial configurations of scar tissue representing various injury patterns. The anatomical information for these heart meshes were derived from real subject hearts, sourced from \cite{dawoud2009noninvasive} and anonymized \href{https://drive.google.com/drive/folders/1zRNoRxlwReiVrUJ0MOvjfwoLPgAK5n9T?usp=sharing}{here}. Each configuration was treated as a separate subject. For the task sequence, we selected 9 subjects (taking 3 scars from each heart) in random order, yielding a total of 517 unique simulations. The remaining three scar configurations, totaling 41 simulations, were held out to evaluate generalization performance following continual learning.

On each subject, we simulated macroscopic spatiotemporal propagation sequences of action-potentials by the Aliev-Panfilov model \citep{aliev1996simple} considering approximately $57\pm11$ different origins of activation separated into a 80-20 train-test split. To obtain the heart-surface measurements in the form of extracellular potential from volumetric action potential, we obtained the forward operator by solving Poisson's equation using the coupled mesh-free method and boundary element methods as described in \citep{wang2009physiological,liu2003meshfree,brebbia2012boundary}. Because the action potential simulated by the Aliev-Panfilov model was unit-less in both amplitude and time, the generated signals on the heart were also unit-less. Specifically, in our experiments, we considered the depolarization process that was down-sampled in time and represented by 55 discrete time steps. The synthetic simulation data for the heart meshes is made available upon acceptance.

\subsubsection{Baselines}
To the best of our knowledge, there is no existing work that considers continual learning of cardiac simulation models. As such, there are no direct baselines to compare against. Instead, we adapted relevant stationary models as best as possible to evaluate the contribution of CoMetaPNS.

\paragraph{\textbf{Non-learning-based personalization methods}}
We included classic non-learning baselines that apply iterative optimization routines to estimate the parameters of a cardiac simulation model. In specific, we considered the published work where \textit{tissue excitability} within the Aliev-Panfilov model was estimated by derivative-free Bayesian optimization \citep{dhamala2018high}. We included both optimization formulations discussed in \citep{dhamala2018high}: one parameterizing tissue excitability using a seven-segment division of the cardiac mesh (FS-BO), and another leveraging a variational autoencoder (VAE) to parameterize spatially-varying tissue excitability (VAE-BO). 
Since these methods were designed to optimize a single subject-specific simulation model at a time, for each of the 9 subjects, we optimized the Aliev-Panfilov model using the 
same context data used to adapt CoMetaPNS.
The resulting Aliev-Panfilov model was used to simulate/predict the action potential sequence using other origins of activations in the query set.

\paragraph{\textbf{Learning-based neural Surrogates}} While neural surrogates for cardiac simulations have been explored \citep{cantwell2019rethinking,kashtanova2021ep,fresca2021pod}, existing approaches are typically limited to 2D models on image grids or focus on atrial structures \citep{fresca2021pod}. These models, once trained, remain general and require additional optimization to personalize predictions for subject-specific data. To more directly evaluate the contributions of our meta-model against traditional neural surrogates, we implemented a version of CoMetaPNS without the meta-model or continual learning - resulting in a standard generative model \( p(x_{1:T} |s,c) \), where the embedding \( c \) is inferred directly from \( y_{1:T} \) via \( q(c | y_{1:T}) \). We refer to this baseline as PNS. This comparison allows us to isolate and assess the impact of the meta-learning and continual learning components individually.

\paragraph{\textbf{Alternative CML baselines}} \label{paragraph:metalearning}
To benchmark against traditional CML approaches, we include a gradient-based variant of our meta-model, denoted as MAML-PNS. This model adopts the standard MAML framework, where the shared initialization parameters correspond to the learnable weights of the GRU-based transition function $\mathcal{T}$. Specifically, it optimizes over the parameter set $\{W^*_i, b^*_i, \alpha^*_i, \beta^*_i, \gamma^*_i\}_{i=1}^3$, which are updated through a small number of gradient descent steps using the available context data for each task.

\subsubsection{Evaluation Settings \& Metrics}
To assess the effect of non-stationary subject distributions on CoMetaPNS and its baselines,
we evaluated PNS, MAML-PNS, and CoMetaPNS under three experience replay settings: 
\textit{i)} Naive Learning (NL), where no past samples are replayed; \textit{ii}) Exact Replay (ER), in which the reservoir can accommodate all past samples; and \textit{iii}) Task-Aware Meta-Learning, as introduced in Section~\ref{sec:CatastrophicForgetting}. 
For CoMetaPNS, we further delved into a comparison between Task-Relation Meta-Learning and Task-Aware Meta-Learning strategies. Since the non-learning FS-BO and VAE-BO baselines performed optimization for each subject instance independently, they were not affected by this setting of data streams.

The accuracy of the solutions in forecasting was measured by the mean square error (MSE), spatial correlation coefficient (SCC), and temporal correlation coefficient (TCC) between the reconstructed and actual potential sequence on the heart surface. 
In addition, we considered the dice coefficient (DC) of the abnormal tissue region obtained by thresholding signals with Otsu’s method \citep{otsu1979threshold}.

To evaluate the ability of a model to learn continually, we followed \citep{MER} and examined the above forecasting metrics in two perspectives: \textit{Retained Performance} (RP) and \textit{Learning Performance} (LP). The RP metric represented the average performance across all tasks after they were sequentially considered, emphasizing a model's ability to retain performance on previous tasks. The LP metrics reflected the average performance on a task immediately after it was learned, measuring a model's effectiveness in incorporating new information. To assess the extent of catastrophic forgetting, we also reported the average difference between LP and RP metrics, referred to as \textit{Backward Transfer and Interference} (BTI), with negative values indicating increased forgetting \citep{MER}.

When comparing the choice of the meta-learner in Section~\ref{subsec:computationalcost}, we included two additional metrics related to computational efficiency. \textit{Time-to-Adapt-N} (TTA-N) represented the average time it takes, in seconds, for a model to adapt to its context data in the presence of $N$ tasks. \textit{Time-to-Train} (TTT) represented the average time it takes, in minutes, for a model to train over a sequence of tasks. 

\subsubsection{Other Implementation Details} 
All experiments were run on NVIDIA Tesla T4 GPUs with 16GB memory in instanced cloud systems to control hardware purity. We used PyTorch 1.13.1 and scikit-learn 1.4.2 for deep learning optimization and GMM fitting, respectively.

All models were trained to forecast the $55$ timesteps using only the one-hot mesh encoding of the activation origin. We set the window of \textit{local stationarity} to $2000$ iterations, where each iteration is composed of $5$ active task samples. 
For CL, we used a reservoir size that can accommodate approximately $45\%$ of the total data at any given time, resulting in a reservoir size of $200$. The size of the $k$-shot context set was set to vary, where at each iteration we sample $k$ between 1-5. To ensure fairness, comparable model components between the meta-learners were scaled to maintain consistent total parameter counts of approximately 1.1 million trainable parameters. Shared backbone components had identical hyper-parameters, while model-specific hyper-parameters were tuned over the same amount of GPU hours.

To test LP and RP metrics, we considered held-out testing sets of every dynamics. To test LP, whenever a subject was finished training for the first time, it was evaluated on its testing set for its performance. At the end of the task sequence, on the resulting model, we evaluated it on every subject independently and averaged their performances to get the RP metric. BTI, then, was the average difference between all subjects' LP and their respective RP.

\subsection{The Effect of Continual Learning} 
\label{subsec:naive}
In this section, we examine the necessity of employing suitable continual learning strategies to address the challenges posed by non-stationary cardiac dynamics.

\subsubsection{Quantitative Results} 
\begin{figure*}[!t]
    \begin{center}
        \includegraphics[width=\textwidth]{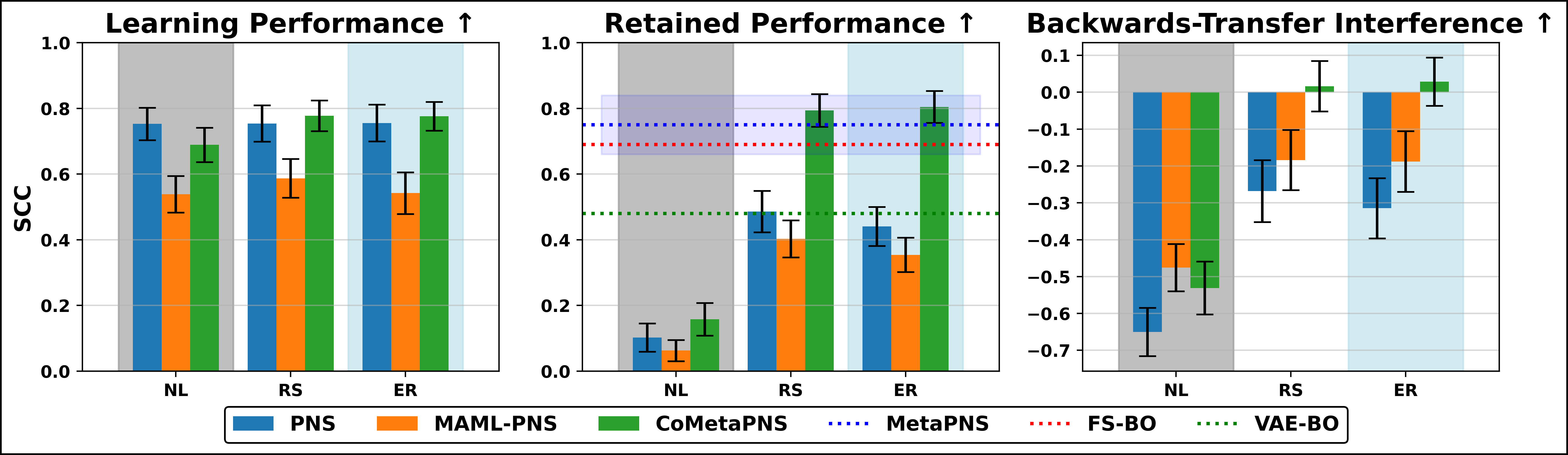}
        \caption{Spatial correlation coefficient (SCC) performance comparison on the synthetic data for the considered baselines across the continual strategies. Horizontal dotted lines represent the average performance of the per-subject baselines. All methods were run over 5 shared seeds.}
        \label{fig:CLBaselinesSCC}
    \end{center}
\end{figure*}

We present results on spatial correlation coefficient (SCC) for the synthetic heart dataset in Fig.~\ref{fig:CLBaselinesSCC}. Gray and blue shaded regions correspond to worst- and best-case scenarios, respectively, representing models trained without (NL) or with full (ER) access to prior task examples. Red (VAE-BO) and green (FS-BO) dotted lines mark the average performance of classic optimization methods personalized to each heart, serving as a reference for RP performance. The blue dotted line represents the meta-model trained in a full stationary setting with access to the full data distribution, termed MetaPNS.

\begin{figure*}
    \begin{center}
        \includegraphics[width=\textwidth]{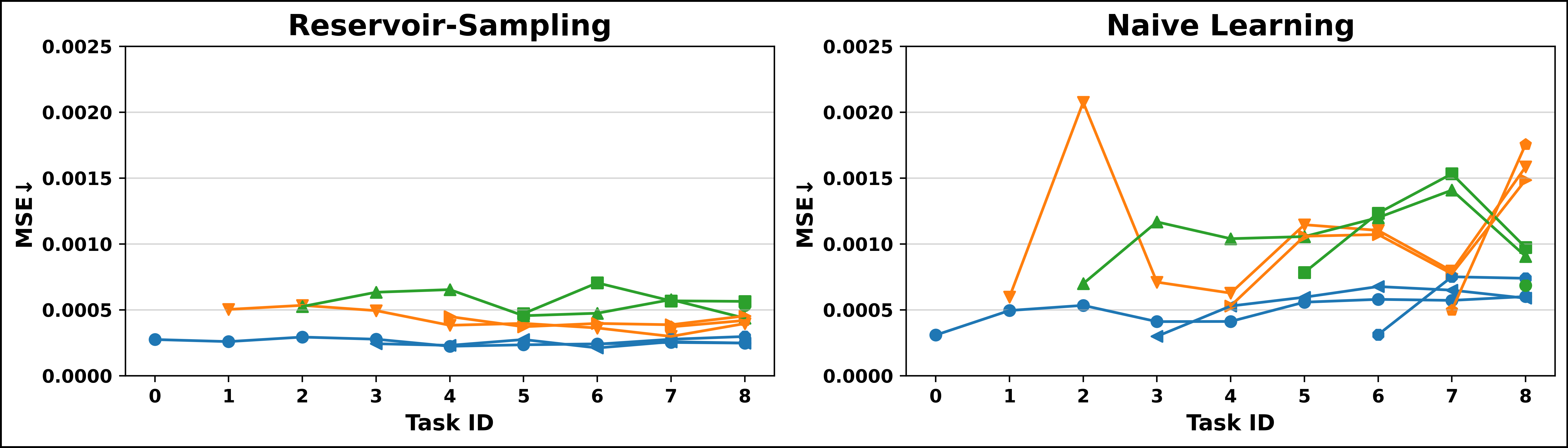}
        \caption{CoMetaPNS with Task-Aware Meta-Learning (left) overcomes catastrophic forgetting in the continual setting vs. naive training (right).}
        \label{fig:NaiveCLComparison}
    \end{center}
\end{figure*}

For LP, all models showed relatively stable performance across experience replay settings, suggesting minimal impact of replay on learning the active task. Interestingly, CoMetaPNS exhibited slightly better LP under the naive setting, potentially due to fully utilizing model capacity on the current subject without the influence of replay. In contrast, RP results revealed that all models suffered from performance degradation when replay was absent, indicating severe forgetting. Introducing a memory mechanism - whether through exact replay or task-aware sampling - consistently mitigated this issue and, for CoMetaPNS, even display positive forward transfer on early tasks. Among all, the proposed combination of a meta-model and task-aware meta-learning provided the most substantial gains, highlighting the necessity of integrating both continual and meta-learning components. Fig.~\ref{fig:NaiveCLComparison} further illustrates per-task performance for CoMetaPNS under settings with (left) and without (right) task-aware meta-learning strategies.

Overall, we can see that CoMetaPNS consistently outperformed all baselines considered including the classic optimization approaches,
and matched the proposed meta-model trained in a stationary setting.

\begin{figure*}[!t]
    \begin{center}
        \includegraphics[width=\textwidth]{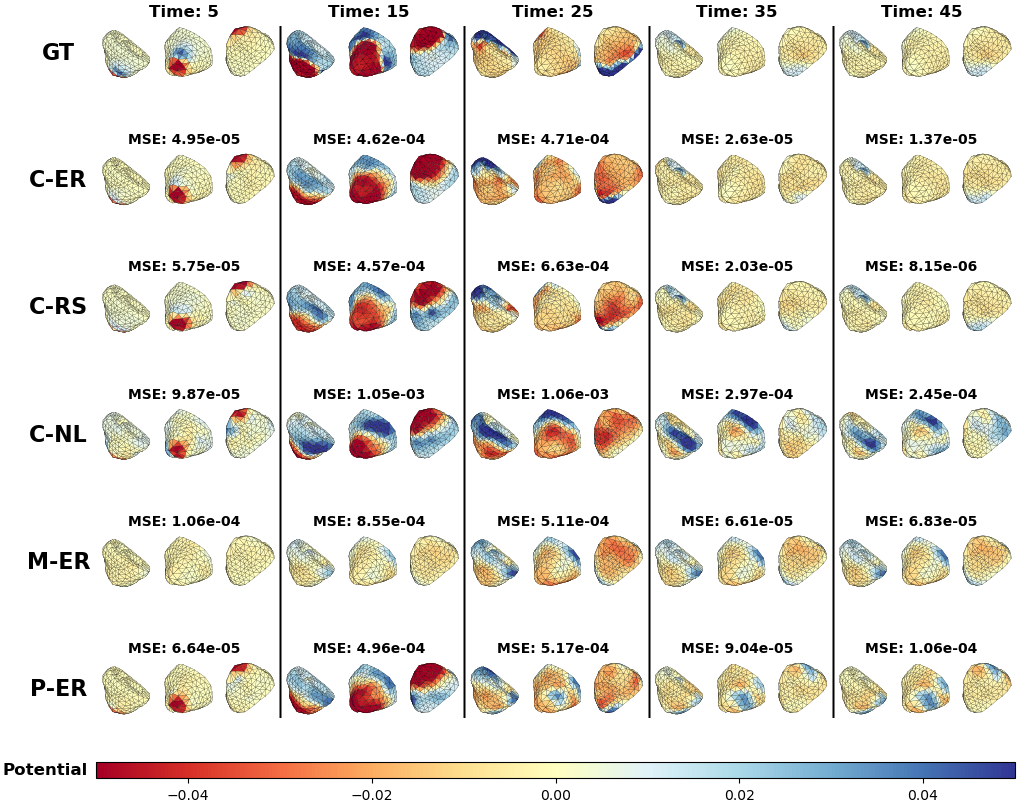}
        \caption{Visual examples of reconstructed electrical activity from the proposed continual meta-model when using continual strategies. Both epicardial and endocardial surfaces are presented. The color bar shows the scaled range of the signal since it is unitless in synthetic data. The MSE value is shown for each model at each timestep. C: CoMetaPNS, M: MAML-PNS, P: PNS. ER: Exact-Replay, RS: Reservoir Sampling, NL: Naive Learning.}
        \label{fig:CardiacMesh}
    \end{center}
    \vspace{-.3cm}
\end{figure*}

\subsubsection{Visual Results} Fig.~\ref{fig:CardiacMesh} provides visual examples of the forecasted electrical activity from the proposed meta-model under each continual strategy, compared against the ground truth. After completing training on the final task, the meta-models were evaluated on the first task when it reappeared at meta-test time. Both reservoir sampling and exact-replay strategies produced accurate forecasting of the propagation dynamics, demonstrating effective mitigation of catastrophic forgetting. Notably, due to variability in subject data availability, exact-replay required sampling uniformly across task IDs rather than the reservoir itself to prevent degraded performance on underrepresented tasks. In contrast, the naive strategy failed to recover the subject-specific tissue properties, resulting in poor reconstruction performance.

For the PNS baselines, we observe that while they effectively capture the initial propagation of activation, they fail to generate smooth or accurate dynamics in the latter half of depolarization. We attribute this to their ability to learn a generic representation of propagation dynamics, lacking the flexibility to adapt to subject-specific features. In contrast, MAML demonstrates more coherent long-term forecasting in terms of per-node activation patterns, but its predictions exhibit notably weaker activation strength. We attribute this to the inherent challenge of simultaneously adapting both spatial and temporal dynamics from a shared initialization in this domain.

\begin{figure*}[!t]
    \begin{center}
        \includegraphics[width=\textwidth]{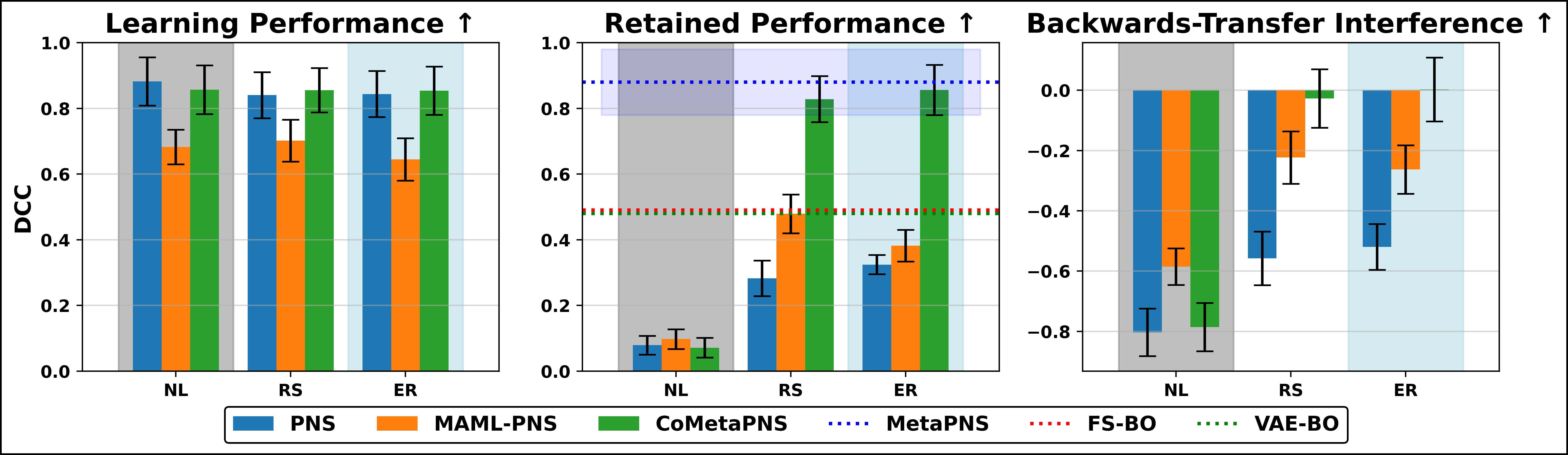}
        \caption{Dice coefficient (DC) performance comparison on the synthetic data for the considered baselines across the continual strategies. Horizontal dotted lines represent the average performance of the per-subject baselines. All methods were run over 5 shared seeds.}
        \label{fig:CLBaselinesDC}
    \end{center}
\end{figure*}

\subsection{The Effect of Meta-Learning} \label{subsec:metalearning}
In this section, we investigate 1) the necessity of the meta-model in learning across heterogeneous cardiac dynamics (by comparing with the PNS baseline without meta-learning), and 2) the impact of different algorithmic priors used for meta-adaptation (by comparing with the MAML-PNS baseline that uses MAML as the meta-learner).

\subsubsection{Quantitative Results}
Fig.~\ref{fig:CLBaselinesDC} presents Dice coefficient (DC) results on the synthetic heart dataset, with full metric comparisons in Table~\ref{tab:Generalization}. A clear performance gap emerges between non-meta and meta-learning approaches: the non-meta PNS baseline fails to consistently capture subject-specific features, while both meta-learners demonstrate improved localization and forecasting. Comparing the two meta-learning strategies, feed-forward CoMetaPNS outperforms MAML-PNS, which only reaches parity with classical optimization methods. This distinction is particularly visible when comparing performance across metrics. Spatial correlation coefficient (SCC) captures overall signal trends over the mesh but is less sensitive to localized boundaries, such as those defining scar tissue. As a result, SCC may overstate the performance of models that align with global signal structure but lack spatial specificity. The Dice coefficient, in contrast, directly evaluates spatial overlap in activation maps and better reflects a model’s ability to recover subject-specific anatomical detail.

Given these observations, we interpret the performance trends of MAML-PNS as indicating an ability to adapt to subject-specific characteristics and represent inactive tissue, but a limitation in capturing global propagation patterns. This aligns with our rationale in Section~\ref{paragraph:metalearning} regarding the difficulty of task adaptation imposed by the gradient-based algorithmic prior in this domain. Conversely, the generic PNS model captures global dynamics effectively through shared parameters across subjects but lacks the capacity to localize subject-specific features due to its static nature. In contrast, the feed-forward conditioning approach demonstrates its advantage through per-vertex conditioning, achieving both strong global dynamics alignment and accurate localization of subject-specific features such as scar regions.

\begin{table}[!t]
\centering
    \caption{Performance metrics of 1) CoMetaPNS, 2) MetaPNS, 3) MAML-PNS, 4) PNS, 5) FS-BO, and 6) VAE-BO during meta-test. For the continual methods, metrics are evaluated on the model after continual training with the Task-Aware Meta-Learning strategy. Classical optimization methods remain fit per-subject as a baseline.}
    \label{tab:Generalization}
    \begin{tabular}{c|c|c|c}
    \toprule
     \midrule
    Model & MSE & SCC & DC \\
    \midrule
    CoMetaPNS & \textbf{4.3$\pm$1.0e-4} & \textbf{0.79$\pm$0.05} & 0.83$\pm$0.07\\
    MetaPNS & 4.5$\pm$1.2e-4 & 0.74$\pm$0.09 & \textbf{0.86$\pm$0.05} \\
    MAML-PNS & 6.7$\pm$0.9e-4 & 0.40$\pm$0.06 & 0.48$\pm$0.06 \\
    PNS & 4.5$\pm$1.4e-4& 0.49$\pm$0.06 & 0.28$\pm$0.05\\
    FS-BO & 5.3$\pm$6.5e-4& 0.69$\pm$0.25& 0.48$\pm$0.34\\
    VAE-BO & 4.8$\pm$2.5e-4& 0.46$\pm$0.15& 0.48$\pm$0.09\\
    \midrule
    \bottomrule
    \end{tabular}
\end{table}

\subsubsection{Visual Results}
\begin{figure*}[!t]
    \begin{center}
        \includegraphics[width=\textwidth]{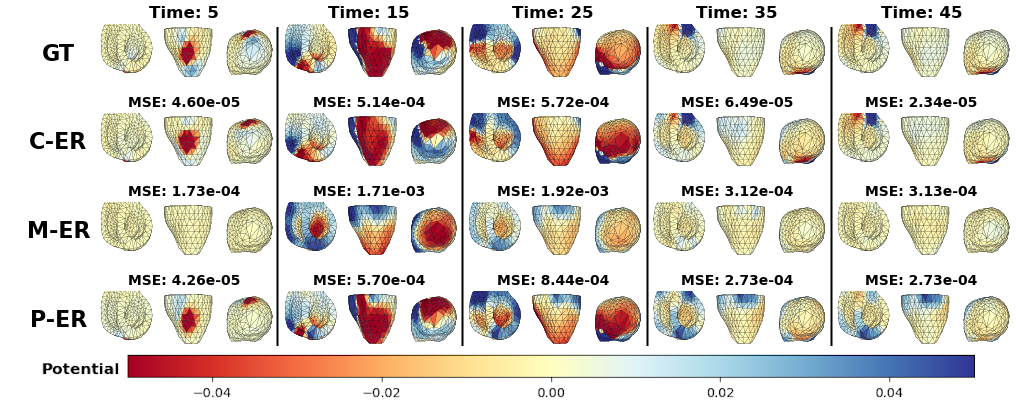}
        \caption{Visual examples of reconstructed electrical activity on both meta-learners considered alongside a non-meta baseline. Both epicardial and endocardial surfaces are presented. The color bar shows the scaled range of the signal since it is unitless in synthetic data. The MSE value is shown for each model at each timestep. C: CoMetaPNS, M: MAML-PNS, P: PNS. ER: Exact-Replay.}
        \label{fig:MetaLearnerComparison}
    \end{center}
\end{figure*}

Fig.~\ref{fig:MetaLearnerComparison} presents visual comparisons of the forecasted electrical activity from the evaluated meta-learners - feed-forward and gradient-based - against the non-meta-learning baseline. Following training on the final task, each meta-model was assessed on the fifth task during meta-testing. The results demonstrate that meta-learning is essential for producing accurate personalized neural surrogates: the non-meta baseline (PNS) fails to recover subject-specific tissue properties and exhibits limited forecasting capability. Furthermore, the selection of the meta-learning algorithm significantly influences the capacity for personalization. The gradient-based MAML model shows weak activation forecasting throughout the sequence and does not sufficiently adapt to subject-specific characteristics in this instance.

\subsubsection{Computational comparisons} \label{subsec:computationalcost}
Table~\ref{tab:CMLCompleteEfficiencyComparison} compares the computational performance of the two meta-learners for model adaptation/personalization, versus classic instance-wise Bayesian optimization as a reference. 
As shown, 
the feed-forward adaptation adopted in MetaPNS substantially outperformed MAML in both 
the time it takes to adapt a model at inference time (TTA) 
and the time it takes to train (TTT), 
across all continual strategies. 
Note that MAML exhibited a linear increase in adaptation time as the number of unique tasks grew. 
In comparison, the feed-forward approach maintained strong efficiency due to its parallelizable inference mechanism. 

When compared to classic optimization baselines, CoMetaPNS achieved similar or better predictive performance at a dramatically lower computational cost: a single Aliev-Panfilov simulation averaged 5 minutes, while the neural surrogate required only approximately $0.71\pm0.05$ seconds. Moreover, Bayesian optimization in the FS-BO required approximately 100 simulation calls per personalization, whereas CoMetaPNS completed the context set embedding in $0.49 \pm 0.05$ seconds on average.

This computational advantage of the feed-forward meta-learner, especially its adaptation efficiency at inference time compared to both classic per-instance optimization and alternative MAML-type of meta-learners, make it appealing for clinical deployment.

\begin{table}
	\caption{Complete adaptation efficiency comparison across baselines. Meta-learning methods were adapted over $100$ batches of data across the synthetic heart dataset, timing each batch independently. Bayes-Opt represents the average per-subject time to personalize for the FS-BO baseline. TTA [X]: Time-to-adapt X tasks. TTT: Time-to-train.}
    \centering
    \begin{tabular}{rl|c}
		\toprule
			Model & Metric & Time  \\
        \midrule
		 \multirow{3}*{Bayes-Opt}      & TTA [1] (seconds)  & $\sim30000$  \\
                                        & TTA [9] (seconds) & $\sim360000$  \\
								      & TTT (minutes)      & $\sim6000$  \\
		\midrule
		 \multirow{3}*{MAML }  & TTA [1] (seconds)  & $1.37\pm0.08$ \\
                                        & TTA [9] (seconds) & $4.17\pm0.12$  \\
								        & TTT (minutes)      & $3789\pm108$ \\
		\midrule
		 \multirow{3}*{Feed-Forward}   & TTA [1] (seconds) & $\mathbf{0.49\pm0.05}$ \\
                                        & TTA [9] (seconds) & $\mathbf{1.48\pm0.07}$  \\
								        & TTT (minutes)     &  $\mathbf{696\pm23}$ \\ 
		\midrule
	\bottomrule
    \end{tabular}
	\label{tab:CMLCompleteEfficiencyComparison}
\end{table}

\subsection{Meta-Model Pre-Training} \label{subsec:metapretraining}
In many clinical deployment scenarios, it is not uncommon to expect an initial stage of data collection and model training prior to model deployment. This opens the possibility that continual learning does not begin entirely from scratch but rather from an initial distribution of subjects. Indeed, some CML works consider this scenario by first pre-training a meta-model on a small stationary subject distribution prior to its use in a continual setting \citep{OSAKA,MOCA}. In this section, we considered CoMetaPNS in this setting and evaluated the effect of this meta-model pre-training on the sample efficiency of CoMetaPNS in continually adapting to new cardiac dynamics.

More specifically, we considered three additional held-out scar-geometry configurations in addition to the nine considered in the previous sections. 
We started with a MetaPNS pretrained a stationary distribution of the nine tasks, and compared
1) its performance if directly applied to the held-out tasks (meta-generalization), 
2) its continual learning over the held-out tasks (CoMetaPNS-pretrained), 
and 3) CoMetaPNS trained from scratch on the same tasks (CoMetaPNS).
For reference, we further included a stationary MetaPNS model trained on all 12 tasks during meta-training. 
To assess the effect of meta-model pre-training on continual learning sample efficiency,
we evaluated the continual learning methods under varying levels of data availability per new task, from 3 to 10 unique simulations. The stationary MetaPNS is trained with full data access, averaging $13.7 \pm 2.5$ samples per task.

\begin{figure*}[!t]
    \begin{center}
        \includegraphics[width=\textwidth]{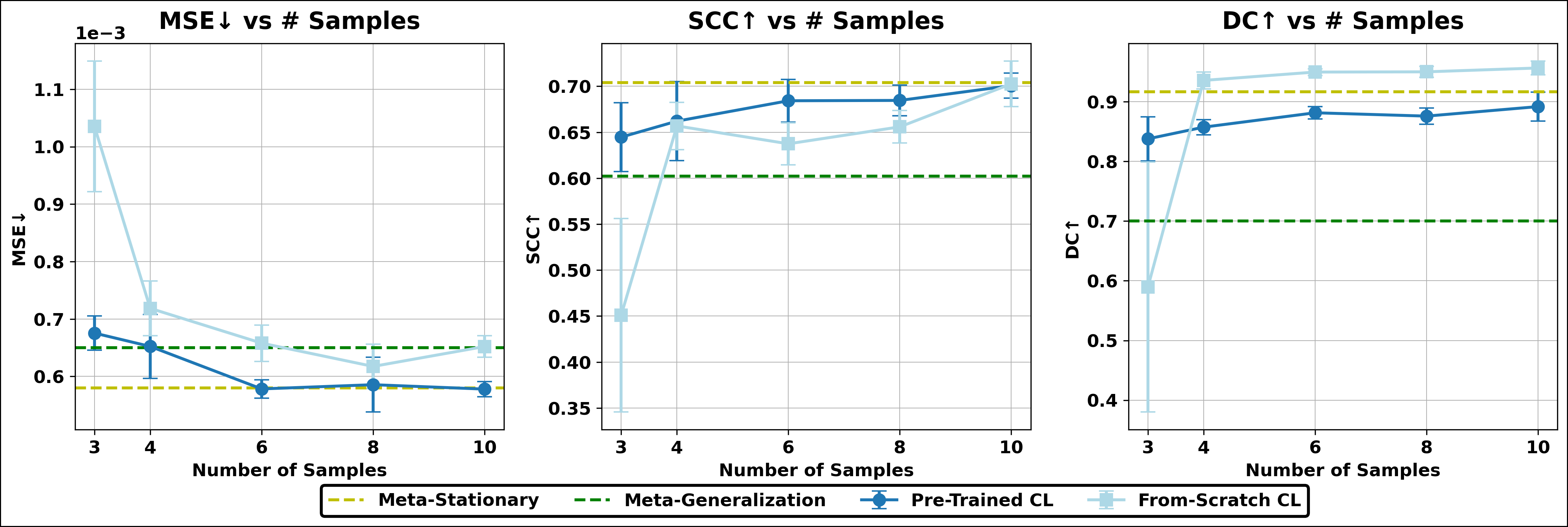}
        \caption{Performance metrics evaluating the effectiveness of using a pre-trained meta-model as initialization for continual learning on held-out tasks when subject to varying levels of per-task sample availability. }
        \label{fig:ContinualVsStationary}
    \end{center}
\end{figure*}

Figure~\ref{fig:ContinualVsStationary} summarizes results across all metrics, with a focus on RP for the continual learning methods. In the low-data regime, CoMetaPNS-pretrained demonstrated a marked improvement over both CoMetaPNS from-scratch and pure meta-generalization, even with just a few samples per task. As data availability increases, both CoMetaPNS-pretrained and CoMetaPNS tend to converge toward the performance of the full stationary model.

More specifically, 
under limited data, CoMetaPNS from-scratch consistently underperformed against the CoMetaPNS-pretrained counterpart in SCC and MSE, indicating challenges in learning general propagation dynamics from sparse observations. The SCC gap narrows with increasing data. 
Curiously, for DC, a notable inversion appears: CoMetaPNS from-scratch surpasses even the full stationary version. We attribute this to an increased capacity of the meta-model to specialize towards the tissue-specific properties of these tasks, though at the cost of reduced propagation fidelity.

\subsection{Additional Benefits of Task-Relational CoMetaPNS}

\subsubsection{
Novel Subject Identification} \label{sec:ClusterAblation}
\begin{figure}[!t]
    \begin{center}
        \includegraphics[width=\columnwidth]{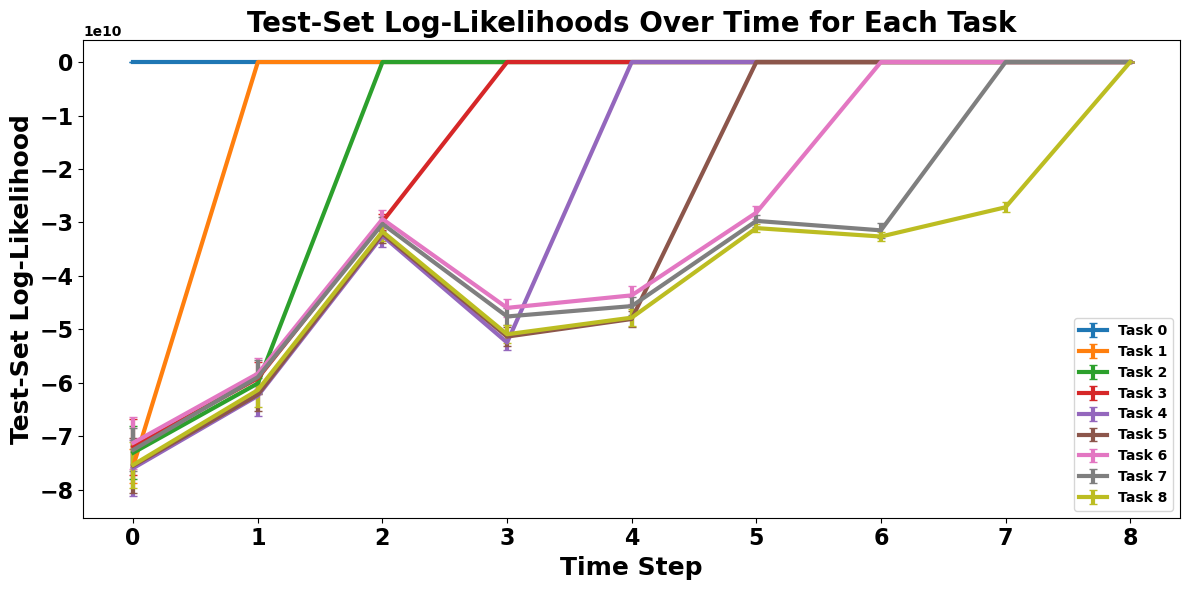}
        \caption{GMM log-likelihoods of all tasks' samples over the sequence, becoming known to the model when its index is in the sequence.}
        \label{fig:ClusterFigure}
    \end{center}
\end{figure}

We further analyzed the benefits of the proposed GMM-based task-relational meta-learning in Fig.~\ref{fig:ClusterFigure}, where the average log-likelihood of each task’s test samples, relative to their nearest cluster mean, shows that the model effectively distinguishes between known and unknown data sources over time. Prior to a new data source’s integration, its samples exhibit consistently low likelihoods across all clusters, while once incorporated, the model reliably re-identifies and clusters future samples from the same source. 

To quantify this benefit of task identification strategy compared to the Task-Aware Meta-Learning strategy, we compared the computational performance of the fast-adaptation meta-model for a given identified known task versus always updating the weights of the meta-model. For the same number of iterations, data, and training structure - with the only difference being whether backpropagation was applied - we see an average speed improvement of $37.7\%$ for processing all iterations. This provides a significant improvement in computational efficiency in the event of recurring, or similar, data distributions in the continual stream where we should not require retraining on known data.

\subsubsection{Clustering Performance}
\begin{figure}[!t]
    \begin{center}
        \includegraphics[width=\columnwidth]{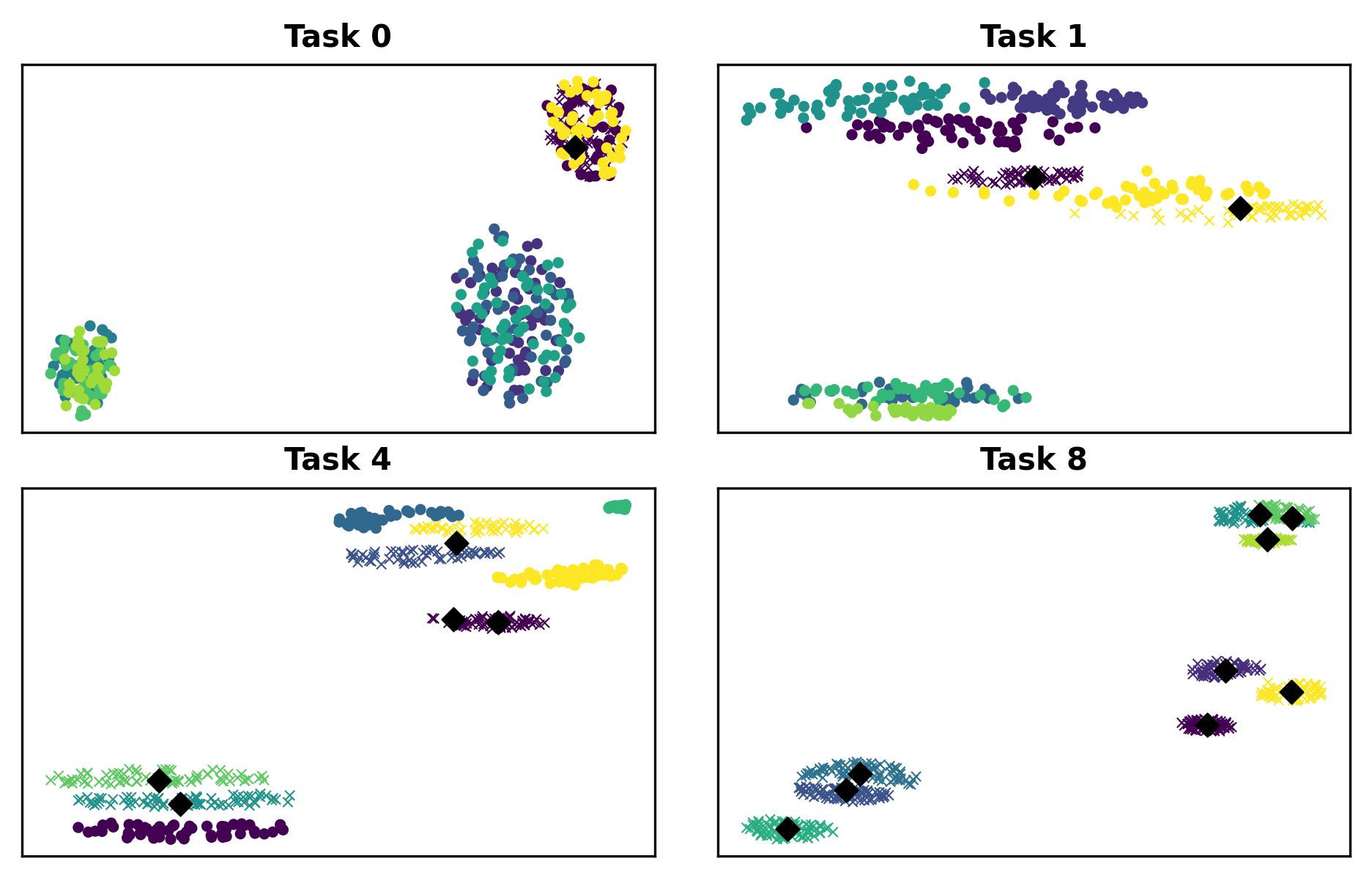}
        \caption{2D t-SNE visualization of the task-relational reservoir over the task sequence. Black stars represent the GMM's cluster means. X-marked samples are known. O-marked samples are unknown.}
        \label{fig:TaskEmeddingSequece}
    \end{center}
\end{figure}

Fig.~\ref{fig:TaskEmeddingSequece} illustrates the trajectory of context embeddings from the task-relational reservoir across the task sequence, along with GMM cluster means, visualized in 2D via t-SNE. To align varying-sized embeddings resulting from different heart mesh geometries, shorter embeddings were padded with random noise to match the largest. While this introduces mesh-specific projection bias, clear intra-mesh clustering by material properties is still observed. It can also be seen that there is clear separation between unknown and known tasks within the embedding space, supporting the use of log-likelihood as a means of novel subject identification.

\begin{figure}[!t]
    \begin{center}
        \includegraphics[width=\columnwidth]{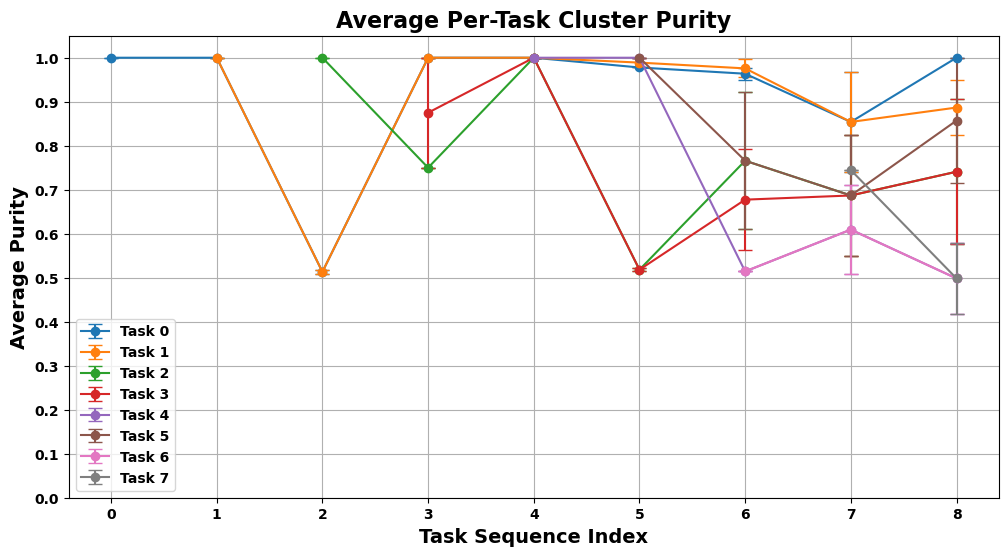}
        \caption{Average cluster purity of each task’s samples over the task sequence. A purity of 1.0 indicates that all samples from a task are assigned to clusters containing only samples from that same task.}
        \label{fig:ClusterPurityFigure}
    \end{center}
\end{figure}

 Given the set of identified known samples, for each underlying task ID, we compute the average purity of that task's meta-embeddings under the GMM clustering. For each task, we identify all clusters containing at least one of its samples and calculate the purity of each, averaging these values to determine the task's clustering purity. Fig.~\ref{fig:ClusterPurityFigure} shows these purities across the task sequence. While the GMM clustering achieves generally stable separation of task embeddings, it is not perfect, with an average final-task purity of $0.74 \pm 0.18$. Importantly, this analysis is based only on samples deemed "known" via log-likelihood thresholds, preserving the method’s ability to identify novel data sources. However, in some cases, the clustering fails to distinguish subtle differences in tissue properties on the same mesh structure. This can be seen in Fig.~\ref{fig:TaskEmeddingSequece} where some of the cluster means across the sequence represent more than one task.

\subsubsection{Effect of Rare Tasks}
\begin{figure}[!t]
    \begin{center}
        \includegraphics[width=\columnwidth]{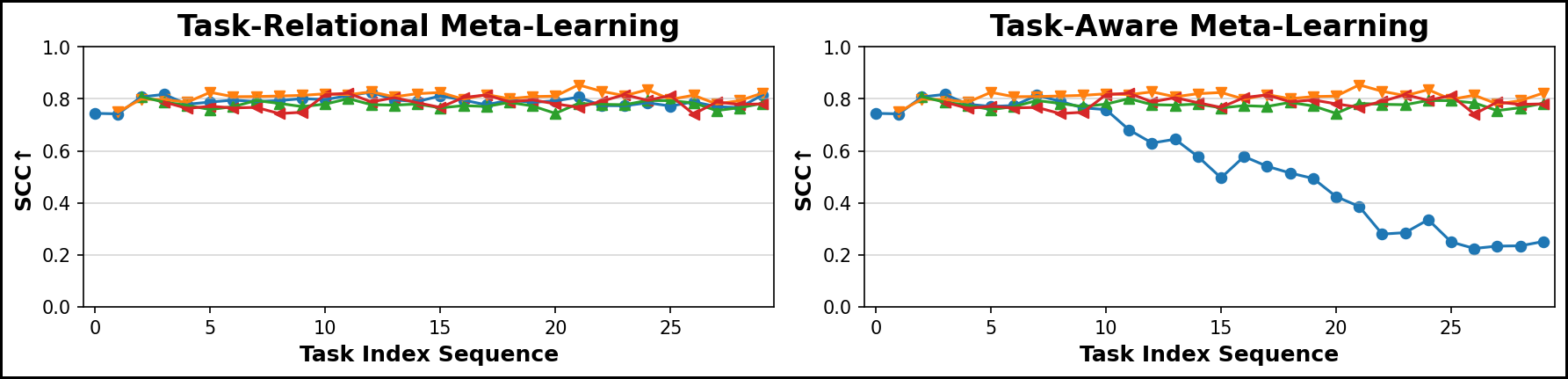}
        \caption{Task-Relational vs. Task-Aware strategies on imbalanced tasks where the first task appears once while others cycle continually. Performance shown on the SCC metric.}
        \label{fig:TaskRelationalAblation}
    \end{center}
\end{figure}

As discussed previously, a key limitation of the Task-Aware Meta-Learning strategy is its assumption that each task corresponds to a distinct subject, which results in a fragmented and imbalanced reservoir as the number of task boundaries increases. To evaluate the impact of this on under-represented tasks, we conducted an ablation experiment with 25 subject-switching boundaries, where one task appeared only at the beginning and three other tasks cycled through the remaining boundaries. The reservoir was restricted to 200 samples. As shown in Fig.~\ref{fig:TaskRelationalAblation}, the SCC performance of each subject over the sequence reveals that the Task-Aware strategy retained only about 5 samples from the first task by the end, leading to severe performance degradation. In contrast, the Task-Relational strategy maintained a more balanced allocation of approximately 50 samples per task, preserving performance even for the initially rare task.

\subsubsection{Computational Requirements} We evaluate the computational requirements of the GMM during training. This strategy introduces additional computational overhead only at task boundaries and does not affect adaptation efficiency at test time. GMM fitting is performed using the scikit-learn library on the CPU, requiring data transfer of meta-embeddings from the GPU, which introduces a minor slowdown. Compared to a meta-model implementation without GMM fitting, this addition resulted in only a $2.5\%$ increase in training time, a modest cost given the benefits of task-relational modeling. Together, these results highlight the embedding space's utility for robust subject-level identification at a limited computational cost.

\section{Real-Data Experiments} \label{sec:realdataexperiments}

\subsection{Data, Baselines, \& Experimental Setup}

Finally, we evaluated CoMetaPNS using \textit{in-vivo} recordings from an animal model experiment \citep{RSM:Ber2021}, sourced from the \href{https://ceg.sci.utah.edu}{Computational Electrocardiology Group}. Cardiac activation sequences were generated via bipolar stimulation using intramural plunge needles at four distinct sites: left ventricular (LV) base, LV apex, LV freewall, and LV septum. Recordings were collected from a 247-electrode epicardial sock array with a spacing of $6.5 \pm 1.3$ mm. Geometric surfaces were reconstructed per recording, and five stimulations were performed at each location, yielding 20 total samples.

 We evaluated CoMetaPNS in two settings: 
 1) direct generalization of CoMetaPNS after it was continually trained on the synthetic data as described in Section \ref{sec:exp:synthetic}, which we refer to as CoMetaPNS-MG (MG for meta-generalization); and 
2) pre-training of MetaPNS on the stationary multi-task synthetic data as described in Section \ref{sec:exp:synthetic}, and then continually fine-tuned on the real data, which we refer to CoMetaPNS-PT (PT for pre-trained). For comparison, we included baselines including the direct generalization of the
 stationary MetaPNS model and the continual, non-meta PNS.

\subsection{Meta-Generalization of CoMetaPNS}
In a challenging meta-generalization setting, we tested the performance of the direct application of CoMetaPNS, MetaPNS, and PNS to real data when trained on synthetic data.
For generalization testing, 
we selected one recording from a stimulation site as the target sample and used one recording from each of the remaining three locations as the context set. Cross-validation was conducted across all valid permutations, ensuring no overlap between target and context samples from the same stimulation site. Model performance on target samples was evaluated using mean squared error (MSE) and spatial correlation coefficient (SCC); Dice coefficient was omitted, as no scar tissue was present in these recordings.

\begin{figure*}[!t]
    \begin{center}
        \includegraphics[width=\textwidth]{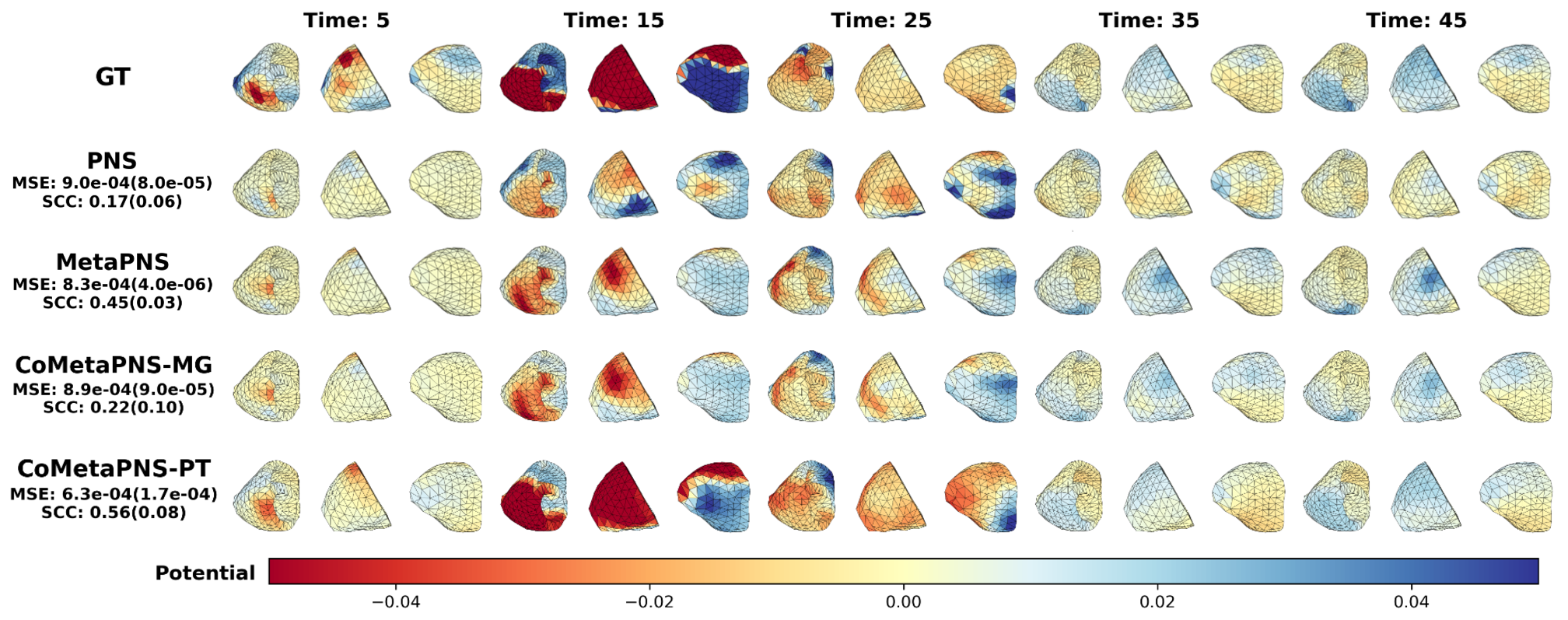}
        \caption{Visual examples of reconstructed electrical activity from the proposed CoMetaPNS when generalizing from synthetic training data to \textit{in-vivo} animal model experiment recordings in addition to continually fine-tuned results. Both epicardial and endocardial surfaces are presented.  PT: Pre-Trained, MG: Meta-Generalization.}
        \label{fig:CardiacMeshReal}
    \end{center}
    \vspace{-.3cm}
\end{figure*}

CoMetaPNS achieved an MSE of 8.9$\pm$0.9e-4 and SCC of 0.22$\pm$0.10, while MetaPNS recorded 8.3$\pm$0.04e-4 and 0.45$\pm$0.03, and PNS achieved 9.0$\pm$0.8e-4 and 0.17$\pm$0.06. Figure~\ref{fig:CardiacMeshReal} illustrates the mesh predictions from all three models. Although there is a notable performance drop compared to synthetic settings, CoMetaPNS continues to generalize better than the non-meta continual PNS. However, its gap to the stationary MetaPNS mirrors the generalization limitations observed in the synthetic experiments (Table~\ref{tab:Generalization}).



\subsection{CoMetaPNS with Pre-Training}
To assess whether pre-training accelerates the incorporation of novel information in real-world scenarios, we conducted additional experiments using the synthetic pre-trained model from Section~\ref{subsec:metapretraining} and adapted it to the \textit{in-vivo} dataset through limited continual fine-tuning. For each experimental trial, one pacing site was designated as the held-out test set, while the remaining three sites constituted the training data. These training sites were partitioned into disjoint context-query sets per-batch and used for meta-learning over a fixed training window. Following this adaptation phase, we extracted a meta-embedding from the training pacing sites and evaluated the model's performance on the held-out site, thereby measuring its capacity to acquire and transfer knowledge to unseen locations. To ensure statistical robustness, we performed leave-one-out cross-validation across all pacing sites and repeated each configuration across 5 random seeds.

Compared to CoMetaPNS-MG when directly applied to the real data, CoMetaPNS-PT after continually fine-tuning on the real data achieved an MSE of 6.3$\pm$1.7e-4 and SCC of 0.56$\pm$0.08. These performance improvements are consistent with our previous findings when leveraging limited training data for rapid adaptation. Figure~\ref{fig:CardiacMeshReal} illustrates the spatial predictions on the cardiac mesh, before and after fine-tuning. These results demonstrate that meta-models pre-trained exclusively on synthetic cardiac data can effectively bridge the domain gap to real cardiac recordings with minimal adaptation, despite the domain gap between computational simulations and \textit{in-vivo} animal recordings.


\section{Conclusion}
In this work, we introduced CoMetaPNS, a continual meta-learning framework designed to learn personalized neural surrogates from non-stationary streams of cardiac data. By combining a feed-forward meta-model with continual experience replay and Bayesian clustering for subject identification, CoMetaPNS enables efficient and scalable personalization without the need for retraining. Empirical results show that maintaining predictive performance across dynamically evolving subject distributions requires the joint use of both meta-learning and continual learning components. This framework provides a practical approach for real-world, personalized cardiac simulation, and future work will explore its extension to diverse observational data types and more complex clinical applications.

This study was conducted extensively on synthetic data with a feasibility study on real animal-model data. This constraint is primarily due to the authors' lack of access to \textit{in-vivo} recordings from 
experimental or clinical subjects undergoing cardiac mapping and stimulation procedures. 
The feasibility study of the ability of CoMetaPNS-PT to quickly adapt to real data using samples from only three pacing sites however provides a glimpse of the significant potential for CoMetaPNS-PT to be used with limited real data.
Future work will focus on this aspect. 
 
Various modeling choices exist for representing cardiac electrical activity in virtual heart simulations, including heart surface potentials \citep{malik2018machine,karoui2019spatial,horvath2019deep} and transmembrane voltages defined over the volumetric mesh of the heart \citep{wang2009physiological,dhamala2019bayesian,ghimire2018generative,ghimire2019improving,ghimire2019noninvasive}. This study focuses on the former, as it represents the most widely adopted formulation and is used in commercial systems. Extending CoMetaPNS to volumetric representations presents additional challenges, particularly in constructing suitable hierarchical graph structures for spatial decoding. Moreover, the significantly larger graph size associated with volumetric models would further increase the computational cost of training.

A complete cardiac signal consists of two primary stages: depolarization and repolarization. However, accurately determining the timing of local electrical activation and recovery from transmembrane voltages or extracellular potentials remains an open challenge. In this study, we restricted our focus to the depolarization stage due to the current limitations of neural networks in capturing more complex physiological dynamics. Future work will explore the use of more advanced neural architectures to better model the full sequence of cardiac electrical activity.

\section{Declaration of generative AI and AI-assisted technologies in the manuscript preparation process.}

During the preparation of this work, the author(s) used Claude Sonnet 4.5 in order to perform writing style and grammar tweaks. No generative tools were used for research purposes, literature review, or content generation. After using this tool/service, the author(s) reviewed and edited the content as needed and take(s) full responsibility for the content of the published article.

\section*{Acknowledgments}
We thank the \href{https://ceg.sci.utah.edu}{Computational Electrocardiology Group} for their support in providing access to and resources for the \textit{in-vivo} animal heart data used within Section~\ref{sec:realdataexperiments}. This study was supported by National Institutes of Health grant 1R01HL145590, 2R01HL145590 and R01NR018301, and National Science Foundation grant number OAC-2212548. 

\bibliographystyle{elsarticle-harv}
\bibliography{biblio}

@article{Kasim20published,
	Author = {M.~F. Kasim and D. Watson-Parris and L. Deaconu and S. Oliver and H. Hatfield and D.~H. Froula and G. Gregori and M. Jarvis and S. Khatiwala and J. Korenaga and J. Topp-Mugglestone and E. Viezzer and S.~M. Vinko},
	Journal = {Machine Learning: Science and Technology},
	Title = {Building high accuracy emulators for scientific simulations with deep
neural architecture search},
    pages = {in press},
	Year = {2021}
}

@article{prakosa2018personalized,
  title={Personalized virtual-heart technology for guiding the ablation of infarct-related ventricular tachycardia},
  author={Prakosa, Adityo and Arevalo, Hermenegild J and Deng, Dongdong and Boyle, Patrick M and Nikolov, Plamen P and Ashikaga, Hiroshi and Blauer, Joshua JE and Ghafoori, Elyar and Park, Carolyn J and Blake, Robert C and others},
  journal={Nature biomedical engineering},
  volume={2},
  number={10},
  pages={732--740},
  year={2018},
  publisher={Nature Publishing Group}
}

@article{arevalo2016arrhythmia,
  title={Arrhythmia risk stratification of patients after myocardial infarction using personalized heart models},
  author={Arevalo, Hermenegild J and Vadakkumpadan, Fijoy and Guallar, Eliseo and Jebb, Alexander and Malamas, Peter and Wu, Katherine C and Trayanova, Natalia A},
  journal={Nature communications},
  volume={7},
  number={1},
  pages={1--8},
  year={2016},
  publisher={Nature Publishing Group}
}

@article{aliev1996simple,
  title={A simple two-variable model of cardiac excitation},
  author={Aliev, Rubin R and Panfilov, Alexander V},
  journal={Chaos, Solitons \& Fractals},
  volume={7},
  number={3},
  pages={293--301},
  year={1996},
  publisher={Elsevier}
}

@article{niederer2020creation,
  title={Creation and application of virtual patient cohorts of heart models},
  author={Niederer, SA and Aboelkassem, Yasser and Cantwell, Chris D and Corrado, Cesare and Coveney, Sam and Cherry, Elizabeth M and Delhaas, Tammo and Fenton, Flavio H and Panfilov, AV and Pathmanathan, P and others},
  journal={Philosophical Transactions of the Royal Society A},
  volume={378},
  number={2173},
  pages={20190558},
  year={2020},
  publisher={The Royal Society Publishing}
}

@article{cantwell2019rethinking,
  title={Rethinking multiscale cardiac electrophysiology with machine learning and predictive modelling},
  author={Cantwell, Chris D and Mohamied, Yumnah and Tzortzis, Konstantinos N and Garasto, Stef and Houston, Charles and Chowdhury, Rasheda A and Ng, Fu Siong and Bharath, Anil A and Peters, Nicholas S},
  journal={Computers in biology and medicine},
  volume={104},
  pages={339--351},
  year={2019},
  publisher={Elsevier}
}

@inproceedings{kashtanova2021ep,
  title={Ep-net 2.0: out-of-domain generalisation for deep learning models of cardiac electrophysiology},
  author={Kashtanova, Victoriya and Ayed, Ibrahim and Cedilnik, Nicolas and Gallinari, Patrick and Sermesant, Maxime},
  booktitle={International Conference on Functional Imaging and Modeling of the Heart},
  pages={482--492},
  year={2021},
  organization={Springer}
}

@article{miller2021implementation,
  title={An Implementation of Patient-Specific Biventricular Mechanics Simulations With a Deep Learning and Computational Pipeline},
  author={Miller, Renee and Kerfoot, Eric and Mauger, Charl{\`e}ne and Ismail, Tevfik F and Young, Alistair A and Nordsletten, David A},
  journal={Frontiers in physiology},
  volume={12},
  pages={1398},
  year={2021},
  publisher={Frontiers}
}

@article{fresca2021pod,
  title={POD-enhanced deep learning-based reduced order models for the real-time simulation of cardiac electrophysiology in the left atrium},
  author={Fresca, Stefania and Manzoni, Andrea and Ded{\`e}, Luca and Quarteroni, Alfio},
  journal={Frontiers in physiology},
  volume={12},
  pages={1431},
  year={2021},
  publisher={Frontiers}
}

@article{dhamala2018quantifying,
  title={Quantifying the uncertainty in model parameters using Gaussian process-based Markov chain Monte Carlo in cardiac electrophysiology},
  author={Dhamala, Jwala and Arevalo, Hermenegild J and Sapp, John and Hor{\'a}cek, B Milan and Wu, Katherine C and Trayanova, Natalia A and Wang, Linwei},
  journal={Medical image analysis},
  volume={48},
  pages={43--57},
  year={2018},
  publisher={Elsevier}
}

@article{wong2015velocity,
  title={Velocity-based cardiac contractility personalization from images using derivative-free optimization},
  author={Wong, Ken CL and Sermesant, Maxime and Rhode, Kawal and Ginks, Matthew and Rinaldi, C Aldo and Razavi, Reza and Delingette, Herv{\'e} and Ayache, Nicholas},
  journal={Journal of the mechanical behavior of biomedical materials},
  volume={43},
  pages={35--52},
  year={2015},
  publisher={Elsevier}
}

@article{sermesant2012patient,
  title={Patient-specific electromechanical models of the heart for the prediction of pacing acute effects in CRT: a preliminary clinical validation},
  author={Sermesant, Maxime and Chabiniok, Radomir and Chinchapatnam, Phani and Mansi, Tommaso and Billet, Florence and Moireau, Philippe and Peyrat, Jean-Marc and Wong, K and Relan, Jatin and Rhode, Kawal and others},
  journal={Medical image analysis},
  volume={16},
  number={1},
  pages={201--215},
  year={2012},
  publisher={Elsevier}
}

@inproceedings{zettinig2013fast,
  title={Fast data-driven calibration of a cardiac electrophysiology model from images and ECG},
  author={Zettinig, Oliver and Mansi, Tommaso and Georgescu, Bogdan and Kayvanpour, Elham and Sedaghat-Hamedani, Farbod and Amr, Ali and Haas, Jan and Steen, Henning and Meder, Benjamin and Katus, Hugo and others},
  booktitle={International Conference on Medical Image Computing and Computer-Assisted Intervention},
  pages={1--8},
  year={2013},
  organization={Springer}
}

@article{dhamala2020embedding,
  title={Embedding high-dimensional bayesian optimization via generative modeling: parameter personalization of cardiac electrophysiological models},
  author={Dhamala, Jwala and Bajracharya, Pradeep and Arevalo, Hermenegild J and Sapp, John L and Hor{\'a}cek, B Milan and Wu, Katherine C and Trayanova, Natalia A and Wang, Linwei},
  journal={Medical image analysis},
  volume={62},
  pages={101670},
  year={2020},
  publisher={Elsevier}
}

@inproceedings{dhamala2018high,
  title={High-dimensional bayesian optimization of personalized cardiac model parameters via an embedded generative model},
  author={Dhamala, Jwala and Ghimire, Sandesh and Sapp, John L and Hor{\'a}{\v{c}}ek, B Milan and Wang, Linwei},
  booktitle={International Conference on Medical Image Computing and Computer-Assisted Intervention},
  pages={499--507},
  year={2018},
  organization={Springer}
}

@article{coveney2021bayesian,
  title={Bayesian Calibration of Electrophysiology Models Using Restitution Curve Emulators},
  author={Coveney, Sam and Corrado, Cesare and Oakley, Jeremy E and Wilkinson, Richard D and Niederer, Steven A and Clayton, Richard H},
  journal={Frontiers in Physiology},
  volume={12},
  pages={1120},
  year={2021},
  publisher={Frontiers}
}

@incollection{neumann2020machine,
  title={Machine learning methods for robust parameter estimation},
  author={Neumann, Dominik and Mansi, Tommaso},
  booktitle={Artificial Intelligence for Computational Modeling of the Heart},
  pages={161--181},
  year={2020},
  publisher={Elsevier}
}

@article{giffard2016noninvasive,
  title={Noninvasive personalization of a cardiac electrophysiology model from body surface potential mapping},
  author={Giffard-Roisin, Sophie and Jackson, Thomas and Fovargue, Lauren and Lee, Jack and Delingette, Herv{\'e} and Razavi, Reza and Ayache, Nicholas and Sermesant, Maxime},
  journal={IEEE Transactions on Biomedical Engineering},
  volume={64},
  number={9},
  pages={2206--2218},
  year={2016},
  publisher={IEEE}
}

@inproceedings{finn2017model,
  title={Model-agnostic meta-learning for fast adaptation of deep networks},
  author={Finn, Chelsea and Abbeel, Pieter and Levine, Sergey},
  booktitle={International conference on machine learning},
  pages={1126--1135},
  year={2017},
  organization={PMLR}
}

@inproceedings{ravi2016optimization,
  title={Optimization as a model for few-shot learning},
  author={Ravi, Sachin and Larochelle, Hugo},
  booktitle={International conference on learning representations},
  year={2016}
}

@article{kingma2013auto,
  title={Auto-encoding variational bayes},
  author={Kingma, Diederik P and Welling, Max},
  journal={arXiv preprint arXiv:1312.6114},
  year={2013}
}

@INCOLLECTION{cgal:c-tsms-07,
   AUTHOR       = {Fernando Cacciola},
   BOOKTITLE    = {CGAL User and Reference Manual},
   PUBLISHER    = {},
   TITLE        = {Triangulated Surface Mesh Simplification},
   YEAR         = {2007},
   EDITION      = {3.3},
   EDITOR       = {CGAL Editorial Board},
   URL          = {http://www.cgal.org/Manual/3.3/doc_html/cgal_manual/packages.html\#Pkg:SurfaceMeshSimplification}
}

@inproceedings{fey2018splinecnn,
  title={SplineCNN: Fast geometric deep learning with continuous B-spline kernels},
  author={Fey, Matthias and Eric Lenssen, Jan and Weichert, Frank and M{\"u}ller, Heinrich},
  booktitle={The IEEE Conference on Computer Vision and Pattern Recognition (CVPR)},
  pages={869--877},
  year={2018}
}

@article{chung2014empirical,
  title={Empirical evaluation of gated recurrent neural networks on sequence modeling},
  author={Chung, Junyoung and Gulcehre, Caglar and Cho, KyungHyun and Bengio, Yoshua},
  journal={arXiv preprint arXiv:1412.3555},
  year={2014}
}

@inproceedings{jiang2021label,
  title={Label-Free Physics-Informed Image Sequence Reconstruction with Disentangled Spatial-Temporal Modeling},
  author={Jiang, Xiajun and Missel, Ryan and Toloubidokhti, Maryam and Li, Zhiyuan and Gharbia, Omar and Sapp, John L and Wang, Linwei},
  booktitle={International Conference on Medical Image Computing and Computer-Assisted Intervention},
  pages={361--371},
  year={2021},
  organization={Springer}
}

@inproceedings{jiang2020learning,
  title={Learning geometry-dependent and physics-based inverse image reconstruction},
  author={Jiang, Xiajun and Ghimire, Sandesh and Dhamala, Jwala and Li, Zhiyuan and Gyawali, Prashnna Kumar and Wang, Linwei},
  booktitle={International Conference on Medical Image Computing and Computer-Assisted Intervention},
  pages={487--496},
  year={2020},
  organization={Springer}
}

@article{otsu1979threshold,
  title={A threshold selection method from gray-level histograms},
  author={Otsu, Nobuyuki},
  journal={IEEE transactions on systems, man, and cybernetics},
  volume={9},
  number={1},
  pages={62--66},
  year={1979},
  publisher={IEEE}
}

@article{kingma2014adam,
  title={Adam: A method for stochastic optimization},
  author={Kingma, Diederik P and Ba, Jimmy},
  journal={arXiv preprint arXiv:1412.6980},
  year={2014}
}

@Article{RSM:Ber2021,
  author =       "J.A. Bergquist and W.W. Good and B. Zenger and J.D. Tate
                 and L.C. Rupp and R.S. MacLeod",
  title =        "The electrocardiographic forward problem: A benchmark
                 study.",
  journal =      "Comput Biol Med",
  year =         "2021",
  month =        "Jul",
  volume =       "134",
  pages =        "104476",
  pmcid =        "PMC8263490",
}

@inproceedings{MAML,
	title        = {Model-agnostic meta-learning for fast adaptation of deep networks},
	author       = {Finn, Chelsea and Abbeel, Pieter and Levine, Sergey},
	year         = 2017,
	booktitle    = {International conference on machine learning},
	pages        = {1126--1135},
	organization = {PMLR}
}

@article{SNAIL,
  title={A simple neural attentive meta-learner},
  author={Mishra, Nikhil and Rohaninejad, Mostafa and Chen, Xi and Abbeel, Pieter},
  journal={arXiv preprint arXiv:1707.03141},
  year={2017}
}

@article{MER,
  title={Learning to learn without forgetting by maximizing transfer and minimizing interference},
  author={Riemer, Matthew and Cases, Ignacio and Ajemian, Robert and Liu, Miao and Rish, Irina and Tu, Yuhai and Tesauro, Gerald},
  journal={arXiv preprint arXiv:1810.11910},
  year={2018}
}

@article{LAMAML,
  title={La-MAML: Look-ahead Meta Learning for Continual Learning, ML Reproducibility Challenge 2020},
  author={Joseph, Joel and Gu, Alex},
  journal={arXiv preprint arXiv:2102.05824},
  year={2021}
}

@article{MetaBGD,
  title={Task agnostic continual learning via meta learning},
  author={He, Xu and Sygnowski, Jakub and Galashov, Alexandre and Rusu, Andrei A and Teh, Yee Whye and Pascanu, Razvan},
  journal={arXiv preprint arXiv:1906.05201},
  year={2019}
}

@article{OSAKA,
  title={Online fast adaptation and knowledge accumulation (osaka): a new approach to continual learning},
  author={Caccia, Massimo and Rodriguez, Pau and Ostapenko, Oleksiy and Normandin, Fabrice and Lin, Min and Page-Caccia, Lucas and Laradji, Issam Hadj and Rish, Irina and Lacoste, Alexandre and V{\'a}zquez, David and others},
  journal={Advances in Neural Information Processing Systems},
  volume={33},
  pages={16532--16545},
  year={2020}
}

@article{MOCA,
  title={Continuous meta-learning without tasks},
  author={Harrison, James and Sharma, Apoorva and Finn, Chelsea and Pavone, Marco},
  journal={Advances in neural information processing systems},
  volume={33},
  pages={17571--17581},
  year={2020}
}

@inproceedings{jiang2022few,
  title={Few-shot generation of personalized neural surrogates for cardiac simulation via bayesian meta-learning},
  author={Jiang, Xiajun and Li, Zhiyuan and Missel, Ryan and Zaman, Md Shakil and Zenger, Brian and Good, Wilson W and MacLeod, Rob S and Sapp, John L and Wang, Linwei},
  booktitle={International Conference on Medical Image Computing and Computer-Assisted Intervention},
  pages={46--56},
  year={2022},
  organization={Springer}
}

@article{wang2009physiological,
  title={Physiological-model-constrained noninvasive reconstruction of volumetric myocardial transmembrane potentials},
  author={Wang, Linwei and Zhang, Heye and Wong, Ken CL and Liu, Huafeng and Shi, Pengcheng},
  journal={IEEE Transactions on Biomedical Engineering},
  volume={57},
  number={2},
  pages={296--315},
  year={2009},
  publisher={IEEE}
}

@article{ghimire2019noninvasive,
  title={Noninvasive Reconstruction of Transmural Transmembrane Potential With Simultaneous Estimation of Prior Model Error},
  author={Ghimire, Sandesh and Sapp, John L and Hor{\'a}{\v{c}}ek, B Milan and Wang, Linwei},
  journal={IEEE transactions on medical imaging},
  volume={38},
  number={11},
  pages={2582--2595},
  year={2019},
  publisher={IEEE}
}

@inproceedings{malik2018machine,
  title={A machine learning approach to reconstruction of heart surface potentials from body surface potentials},
  author={Malik, Avinash and Peng, Tommy and Trew, Mark L},
  booktitle={2018 40th Annual International Conference of the IEEE Engineering in Medicine and Biology Society (EMBC)},
  pages={4828--4831},
  year={2018},
  organization={IEEE}
}

@inproceedings{karoui2019spatial,
  title={A Spatial Adaptation of the Time Delay Neural Network for Solving ECGI Inverse Problem},
  author={Karoui, Amel and Bendahmane, Mostafa and Zemzemi, Nejib},
  booktitle={International Conference on Functional Imaging and Modeling of the Heart},
  pages={94--102},
  year={2019},
  organization={Springer}
}

@article{horvath2019deep,
  title={Deep learning neural nets for detecting heart activity},
  author={Horvath, Joe and Shien, Lu and Peng, Tommy and Malik, Avinash and Trew, Mark and Bear, Laura},
  journal={arXiv preprint arXiv:1901.09831},
  year={2019}
}

@inproceedings{ghimire2019improving,
  title={Improving generalization of deep networks for inverse reconstruction of image sequences},
  author={Ghimire, Sandesh and Gyawali, Prashnna Kumar and Dhamala, Jwala and Sapp, John L and Horacek, Milan and Wang, Linwei},
  booktitle={International Conference on Information Processing in Medical Imaging},
  pages={153--166},
  year={2019},
  organization={Springer}
}

@inproceedings{dhamala2019bayesian,
  title={Bayesian Optimization on Large Graphs via a Graph Convolutional Generative Model: Application in Cardiac Model Personalization},
  author={Dhamala, Jwala and Ghimire, Sandesh and Sapp, John L and Hor{\'a}{\v{c}}ek, B Milan and Wang, Linwei},
  booktitle={International Conference on Medical Image Computing and Computer-Assisted Intervention},
  pages={458--467},
  year={2019},
  organization={Springer}
}

@inproceedings{ghimire2018generative,
  title={Generative modeling and inverse imaging of cardiac transmembrane potential},
  author={Ghimire, Sandesh and Dhamala, Jwala and Gyawali, Prashnna Kumar and Sapp, John L and Horacek, Milan and Wang, Linwei},
  booktitle={International Conference on Medical Image Computing and Computer-Assisted Intervention},
  pages={508--516},
  year={2018},
  organization={Springer}
}

@book{liu2003meshfree,
author={Liu,G. R.},
year={2003},
title={Mesh free methods: moving beyond the finite element method},
publisher={CRC Press},
address={Boca Raton, Fla},
isbn={0849312388;9780849312380;},
language={English},
}

@book{brebbia2012boundary,
  title={Boundary element techniques: theory and applications in engineering},
  author={Brebbia, Carlos Alberto and Telles, Jos{\'e} Claudio Faria and Wrobel, Luiz C},
  year={2012},
  publisher={Springer Science \& Business Media}
}

@article{trayanova2024computational,
  title={Computational modeling of cardiac electrophysiology and arrhythmogenesis: toward clinical translation},
  author={Trayanova, Natalia A and Lyon, Aurore and Shade, Julie and Heijman, Jordi},
  journal={Physiological reviews},
  volume={104},
  number={3},
  pages={1265--1333},
  year={2024},
  publisher={American Physiological Society Rockville, MD}
}

@article{kolk2024dynamic,
  title={Dynamic prediction of malignant ventricular arrhythmias using neural networks in patients with an implantable cardioverter-defibrillator},
  author={Kolk, Maarten ZH and Ruip{\'e}rez-Campillo, Samuel and Alvarez-Florez, Laura and Deb, Brototo and Bekkers, Erik J and Allaart, Cornelis P and Van Der Lingen, Anne-Lotte CJ and Clopton, Paul and I{\v{s}}gum, Ivana and Wilde, Arthur AM and others},
  journal={EBioMedicine},
  volume={99},
  year={2024},
  publisher={Elsevier}
}

@article{cluitmans2024digital,
  title={Digital twins for cardiac electrophysiology: state of the art and future challenges},
  author={Cluitmans, Matthijs JM and Plank, Gernot and Heijman, Jordi},
  journal={Herzschrittmachertherapie+ Elektrophysiologie},
  volume={35},
  number={2},
  pages={118--123},
  year={2024},
  publisher={Springer}
}

@article{bhagirath2024bits,
  title={From bits to bedside: entering the age of digital twins in cardiac electrophysiology},
  author={Bhagirath, Pranav and Strocchi, Marina and Bishop, Martin J and Boyle, Patrick M and Plank, Gernot},
  journal={Europace},
  volume={26},
  number={12},
  pages={euae295},
  year={2024},
  publisher={Oxford University Press UK}
}

@article{kirkpatrick2017overcoming,
  title={Overcoming catastrophic forgetting in neural networks},
  author={Kirkpatrick, James and Pascanu, Razvan and Rabinowitz, Neil and Veness, Joel and Desjardins, Guillaume and Rusu, Andrei A and Milan, Kieran and Quan, John and Ramalho, Tiago and Grabska-Barwinska, Agnieszka and others},
  journal={Proceedings of the national academy of sciences},
  volume={114},
  number={13},
  pages={3521--3526},
  year={2017},
  publisher={National Academy of Sciences}
}

@article{herrero2022ep,
  title={EP-PINNs: Cardiac electrophysiology characterisation using physics-informed neural networks},
  author={Herrero Martin, Clara and Oved, Alon and Chowdhury, Rasheda A and Ullmann, Elisabeth and Peters, Nicholas S and Bharath, Anil A and Varela, Marta},
  journal={Frontiers in Cardiovascular Medicine},
  volume={8},
  pages={768419},
  year={2022},
  publisher={Frontiers Media SA}
}

@article{gomez2025simulation,
  title={Simulation of parametrized cardiac electrophysiology in three dimensions using physics-informed neural networks},
  author={Gomez, Roshan Antony and St{\"o}cker, Julien and Cans{\i}z, Bar{\i}{\c{s}} and Kaliske, Michael},
  journal={arXiv preprint arXiv:2506.15405},
  year={2025}
}

@article{martinez2025full,
  title={Full-field surrogate modeling of cardiac function encoding geometric variability},
  author={Martinez, Elena and Moscoloni, Beatrice and Salvador, Matteo and Kong, Fanwei and Peirlinck, Mathias and Lesley Marsden, Alison},
  year={2025},
  publisher={arXiv}
}

@article{salvador2024whole,
  title={Whole-heart electromechanical simulations using latent neural ordinary differential equations},
  author={Salvador, Matteo and Strocchi, Marina and Regazzoni, Francesco and Augustin, Christoph M and Dede’, Luca and Niederer, Steven A and Quarteroni, Alfio},
  journal={NPJ Digital Medicine},
  volume={7},
  number={1},
  pages={90},
  year={2024},
  publisher={Nature Publishing Group UK London}
}

@phdthesis{dawoud2009noninvasive,
  title={Noninvasive Imaging of Epicardial Potentials for Clinical Electrophysiology},
  author={Dawoud, Fady},
  year={2009},
  school={Dalhousie University}
}
\end{document}